\documentclass[10pt,twocolumn,letterpaper]{article}
\usepackage[final]{cvpr}
\definecolor{cvprblue}{rgb}{0.21,0.49,0.74}
\usepackage[pagebackref,breaklinks,colorlinks,allcolors=cvprblue,citecolor=cvprblue]{hyperref}
\usepackage{graphicx}
\usepackage{multirow}
\usepackage{booktabs} 
\usepackage{tikz}
\usepackage{float} 
\usepackage{amsmath}
\usepackage{amssymb}
\usepackage{enumitem}
\usepackage{afterpage}
\usepackage{tikz}
\usetikzlibrary{arrows, decorations.markings, arrows.meta, positioning, calc}
\usepackage{overpic}
\usepackage{newtxtext,newtxmath}
\usepackage{helvet}
\usetikzlibrary{calc}
\usepackage[normalem]{ulem}
\usepackage[accsupp]{axessibility} 

\useunder{\uline}{\ul}{}
\newcommand{\Proj}{\mathbf{P}}

\def\confName{CVPR}
\def\confYear{2025}
\title{Multi-person Physics-based Pose Estimation for Combat Sports}

\author{
Hossein Feiz$^{1}$ \quad
David Labbé$^{1}$ \quad
Thomas Romeas$^{2}$ \quad
Jocelyn Faubert$^{2}$ \quad
Sheldon Andrews$^{1}$\\[6pt]
{\small $^{1}$École de technologie supérieure, Montreal, Canada}\\
{\small \tt\small \{hossein.feizollah-zadeh-khoiee.1,david.labbe,sheldon.andrews\}@etsmtl.ca}\\[4pt]
{\small $^{2}$Université de Montréal, Montreal, Canada}\\
{\small \tt\small \{thomas.romeas,jocelyn.faubert\}@umontreal.ca}
}
\newcommand{\teaserfigure}{%
    \begin{center}
        \includegraphics[width=\linewidth]{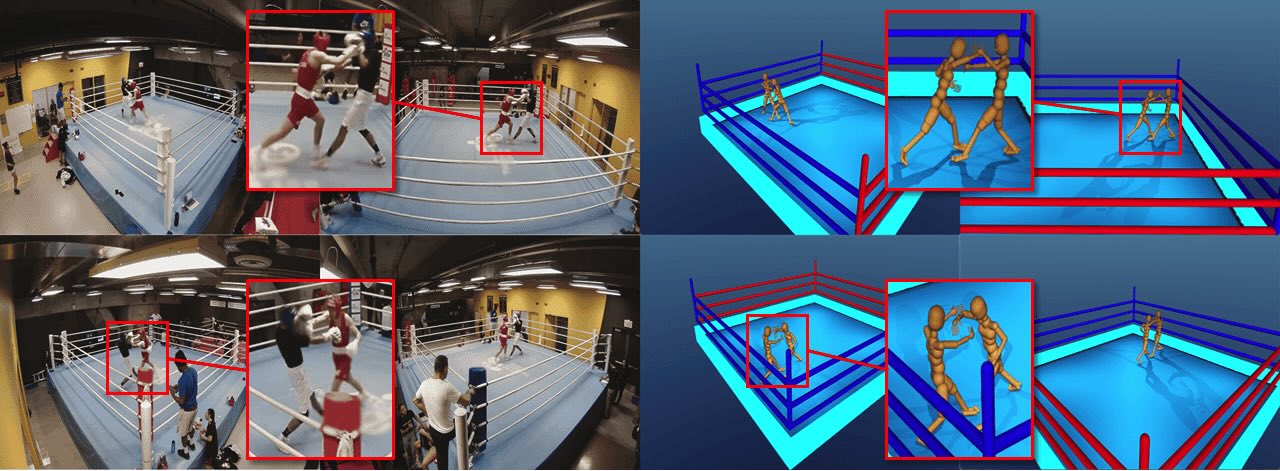}
        \captionsetup{type=figure}
        \captionof{figure}{%
            Multiple camera views of a boxing match (left) are processed by our pipeline to reconstruct the motion of multiple athletes inside a physics-based virtual environment (right). 
        }
        \label{fig:teaser}
    \end{center}
}
\usepackage{balance}

\begin{document}
\twocolumn[{%
    \renewcommand\twocolumn[1][]{#1}%
    \maketitle
    \teaserfigure
}]
\begin{abstract}
We propose a novel framework for accurate 3D human pose estimation in combat sports using sparse multi-camera setups. Our method integrates robust multi-view 2D pose tracking via a transformer-based top-down approach, employing epipolar geometry constraints and long-term video object segmentation for consistent identity tracking across views. Initial 3D poses are obtained through weighted triangulation and spline smoothing, followed by kinematic optimization to refine pose accuracy. We further enhance pose realism and robustness by introducing a multi-person physics-based trajectory optimization step, effectively addressing challenges such as rapid motions, occlusions, and close interactions. Experimental results on diverse datasets, including a new benchmark of elite boxing footage, demonstrate state-of-the-art performance. Additionally, we release comprehensive annotated video datasets to advance future research in multi-person pose estimation for combat sports. Results and dataset can be found on our \href{https://hosseinfeiz.github.io/physpose/}{project page}.
\end{abstract}
\section{Introduction}
\label{sec:intro}

Accurate 3D motion capture in combat sports remains a challenging problem due to rapid and complex athlete interactions, frequent occlusions, and crowded backgrounds. Traditional marker-based optical motion capture systems, while highly precise in controlled environments, become impractical in such dynamic contexts due to calibration issues and marker displacement caused by collisions. Inertial measurement unit (IMU)-based approaches provide freedom of movement but suffer from cumulative positional drift, especially in capturing precise spatial relationships between interacting athletes. Monocular vision-based systems eliminate tracking hardware constraints but typically lack precision and robustness, particularly under frequent occlusions and rapid motion conditions.

To address these challenges, we introduce a novel multi-stage, multi-view framework specifically tailored for combat sports scenarios such as boxing. Our approach effectively integrates robust multi-view 2D pose estimation, leveraging transformer-based architectures and epipolar geometry constraints, with weighted triangulation and spline smoothing for accurate initial 3D joint position estimates. A subsequent kinematic optimization refines these estimates, while our proposed multi-person physics-based trajectory optimization significantly enhances motion realism and resolves artifacts caused by inter-person penetrations and unrealistic interactions.

Experimental results demonstrate that our pipeline successfully reconstructs complex athlete interactions, faithfully adhering to physical constraints such as contact dynamics and collision avoidance. We evaluate our approach using comprehensive multi-view footage of elite boxers (illustrated in Figure~\ref{fig:teaser}) and further validate its accuracy using a supplementary dataset captured via synchronized optical motion capture systems. This supplementary dataset, along with detailed video annotations, will be publicly released to foster further research.

The contributions of our work can be summarized as follows:
\begin{itemize}
\item A robust multi-camera framework integrating kinematic and physics-based optimizations, enabling precise 3D pose estimation from sparse camera setups.
\item A multi-person physics-based trajectory optimization method that significantly reduces inter-person penetration artifacts compared to existing state-of-the-art solutions.
\item Two novel multi-view datasets: (i) Over 20 minutes of video footage featuring elite boxers during diverse, intense sparring scenarios, and (ii) sequences of human interactions accompanied by precise ground truth motions captured with an optical tracking system.
\end{itemize}

\section{Related Work}
\label{sec:related}

\subsection*{Multi-Person 3D Pose Estimation}

Multi-person 3D human pose estimation has advanced significantly, transitioning from traditional two-stage pipelines~\cite{cheng2021topdown} to unified, single-stage methods like ROMP~\cite{sun2021romp} and Decoupled Regression~\cite{jin2022single}, improving efficiency and robustness under occlusion. Transformer-based methods, such as POTR-3D~\cite{park2023robust}, integrate temporal context and cross-person attention, smoothing predictions and resolving depth ambiguities. Weakly supervised approaches, like Ye et al.~\cite{ye2023decoupling}, reconstruct 3D motion from unlabeled web videos, leveraging self-supervised priors. Srivastav \emph{et al.}~\cite{srivastav2024selfpose3d} similarly introduced SelfPose3D, enabling accurate multi-view 3D estimation without requiring explicit 3D labels. Multi-view methods, like PlaneSweepPose~\cite{lin2021multiviewplanesweep}, directly fuse multiple views using plane-sweep stereo for efficient 3D estimation. Wang \emph{et al.}~\cite{wang2021mvp} and Liao \emph{et al.}~\cite{liao2024mvgformer} utilize transformers and geometric reasoning, significantly enhancing cross-view consistency. Zheng \emph{et al.}~\cite{zheng2021poseformer} introduced PoseFormer, showing that transformers effectively capture temporal dependencies, inspiring subsequent multi-person approaches~\cite{park2023robust}. Diffusion models, like DiffPose~\cite{gong2023diffpose} and Diff3DHPE~\cite{zhou2023diff3dhpe}, iteratively refine predictions, modeling ambiguity and ensuring reliable estimation under uncertainty.

\subsection*{Multi-Person Physics-Based 3D Pose Estimation}

Recent physics-based approaches to single- and multi-person 3D pose estimation integrate differentiable physics, biomechanics, and reinforcement learning to mitigate common artifacts such as foot sliding, body inter-penetration, and collisions~\cite{Isogawa2020NLOS,Yuan2021SimPoE,Huang2022MoCon,Yi2022PIP,Gartner2022Trajectory,Gartner2022Differentiable,Tripathi2023IP,Ugrinovic2024MultiPhys,Fieraru2021REMIPS,RempeContactDynamics2020}.
Differentiable physics-based methods explicitly enforce biomechanical constraints by jointly optimizing kinematic and dynamic parameters, ensuring physically realistic motion even under challenging athletic activities~\cite{Gartner2022Trajectory,Gartner2022Differentiable,RempeContactDynamics2020}.
Similarly, reinforcement learning frameworks enable simulated agents to refine pose predictions through physics-informed rewards, reducing artifacts like floating body parts and penetration~\cite{Yuan2021SimPoE,Huang2022MoCon}.
Incorporation of physics modeling has further improved motion plausibility even when using sparse IMU signals as input, particularly by preventing foot sliding and stabilizing the motion~\cite{andrews2016real,Yi2022PIP,Tripathi2023IP}.
Recent advances also address multi-person scenarios by enforcing mutual constraints, ensuring spatial coherence and collision-free motion~\cite{Ugrinovic2024MultiPhys,Fieraru2021REMIPS}.
These methods often benefit from human-scene interaction cues, leveraging ground support and physical laws to promote robust pose estimation in dense crowds.
Overall, the integration of physics-based modeling enhances realism and robustness across diverse applications, establishing a new standard for performance in both single- and multi-person pose reconstruction tasks. 
The incorporation of biomechanical constraints—as shown by Saleem et al.'s BioPose~\cite{saleem2025biopose}—ensures realistic and immersive human motion. Moreover, reinforcement learning approaches, such as the imitation-based method proposed by Peng et al.~\cite{peng2018deepmimic}, contribute to generating physically coherent human animations; collectively, these diverse methodologies offer robust and realistic multi-person motion estimation suitable for sports analysis, crowd dynamics~\cite{RempeContactDynamics2020,Shimada_2020_PhysCap,li2022dd}.

\subsection*{Multi-Person 3D Pose Estimation in Sports and VR Applications}

Multi-person 3D pose estimation in sports poses significant challenges due to fast and unpredictable movements, frequent occlusions, and the need for fine-grained motion capture under high-intensity conditions~\cite{bridgeman2019multi}; early methods relied on controlled environments with multi-camera setups or marker-based motion capture systems that, while accurate, proved costly and inflexible for real-world scenarios. Recent advances have shifted toward leveraging monocular or multi-view video feeds in unconstrained settings to enable applications such as player tracking, action recognition, and performance analysis in sports like basketball, soccer~\cite{chang2024basketball,jiang2024worldpose}, and tennis~\cite{alshami2023pose2trajectory}. To address issues arising from frequent occlusions and complex player interactions, many approaches now integrate spatio-temporal modules or graph-based methods to capture both short-term and long-term dynamics in high-speed sports sequences~\cite{ren2024empowering, sweeting2017discovering}, while Transformer-based architectures have shown promise in effectively modeling global context in dense player formations~\cite{scott2024teamtrack}. The emergence of large-scale sports datasets with rich annotations further facilitates improved generalization under domain-specific constraints~\cite{li2021multisports}, spurring the development of specialized datasets and systems such as \emph{SportsPose}~\cite{ingwersen2023sportspose}, which features markerless motion capture of five high-speed sports validated against commercial marker-based systems, and \emph{AthletePose3D}~\cite{yeung2025athletepose3d}, which captures extreme dynamics across 12 sport motion types and approximately 1.3 million frames to achieve a reduction in pose error by nearly 69\% (from 214~mm to 65~mm). For team sports, \emph{WorldPose}~\cite{jiang2024worldpose} provides a large-scale multi-person dataset from the 2022 FIFA World Cup that captures 3D poses and trajectories of entire soccer teams using a static multi-camera setup, while tailored applications like \emph{Reconstructing NBA Players}~\cite{zhu2020nba} demonstrate the integration of computer vision with domain-specific data to achieve high-resolution 3D mesh reconstruction from single broadcast images despite challenges such as complex poses, motion blur, and occlusions. In dense crowd scenarios, robust multi-view geometry frameworks that combine probabilistic reasoning and cross-view matching have proven effective~\cite{chen2020multi}. In this work, we focus on using multi-person physics-based pose estimation of elite boxers captured during sparring sessions to robustly recover detailed 3D kinematics under challenging occlusion and rapid motion conditions. Our approach uniquely integrates physics simulation with advanced deep learning, enabling precise biomechanical analysis and offering new insights for performance optimization in competitive boxing.
\section{Methodology}
\begin{figure*}[htbp]
  \centering
  \resizebox{\textwidth}{!}{%
    \begin{tikzpicture}
\tikzstyle{process}=[draw,rectangle, rounded corners, fill=yellow!20, minimum width=1cm, auto, on grid, minimum height=1cm, align=center]
\tikzstyle{process4}=[draw,font=\fontsize{8}{4.8}, rectangle, rounded corners,fill=yellow!20, text width=3cm, auto,on grid,align=center]
\tikzstyle{decision}=[draw,font=\fontsize{8}{4.8}, rectangle, rounded corners,fill=red!20, text width=3.7cm, auto,on grid,align=center]
\tikzstyle{process2}=[draw,font=\fontsize{10}{4.8}, rectangle, rounded corners,fill=cyan!20, text width=4cm, auto,on grid,align=center]
\tikzstyle{process3}=[draw, font=\fontsize{10}{4.8}\selectfont,rectangle, rounded corners,fill=cyan!20, rotate=-90, auto,on grid,align=center]

\tikzstyle{vecArrow} = [thick, decoration={markings,mark=at position
   1 with {\arrow[semithick]{open triangle 60}}},
   double distance=1.4pt, shorten >= 5.5pt,
   preaction = {decorate},
   postaction = {draw,line width=1.4pt, white,shorten >= 4.5pt}]
\tikzstyle{vecArrow1} = [thick, decoration={markings,mark=at position
   0.9 with {}},
   double distance=1.4pt, shorten >= 5.5pt,
   preaction = {decorate},
   postaction = {draw,line width=1.4pt, white,shorten >= 4.5pt}]
\tikzstyle{innerWhite} = [semithick, white,line width=1.4pt, shorten >= 4.5pt]
\node[inner sep=0pt] (tri) at (4,-0.35){{\scalebox{.03}{   
\begin{overpic}[scale=5]{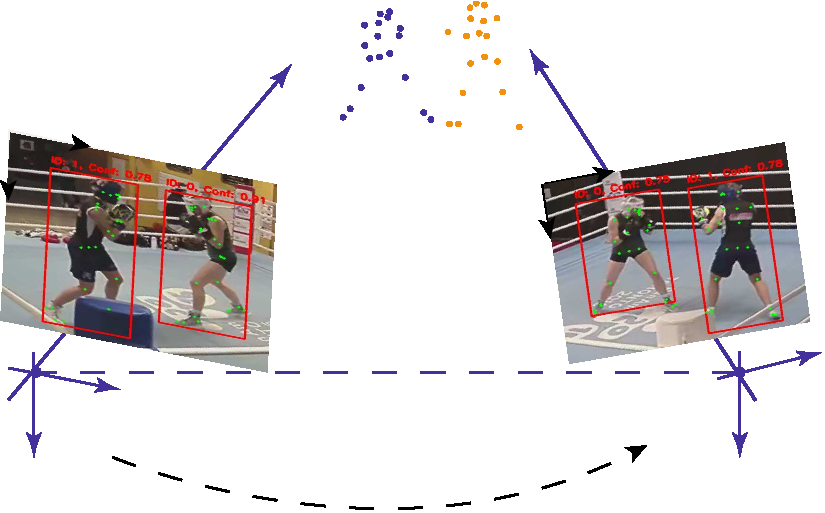}
\put(-2,15){$C_L$}
\put(15,12){$X_L$}
\put(0,5){$Y_L$}
\put(92,14){$C_R$}
\put(91,5){$Y_R$}
\put(102,18){$X_R$}
\put(44,-3){R}
\put(44,13.5){T}
\put(30,9.5){{\fontsize{86}{36}\selectfont triangulation}}
\put(24,47){$Z_L$}
\put(6,47){$u_L$}
\put(-3,39){$v_L$}
\put(50,59){$P_0$}
\put(63,59){$P_1$}
\put(74,47){$Z_R$}
\put(70,42){$u_R$}
\put(62,33){$v_R$}
\end{overpic}}}};
\node[inner sep=0pt] (1) at (-4,0.8){\includegraphics[width=.1\textwidth]{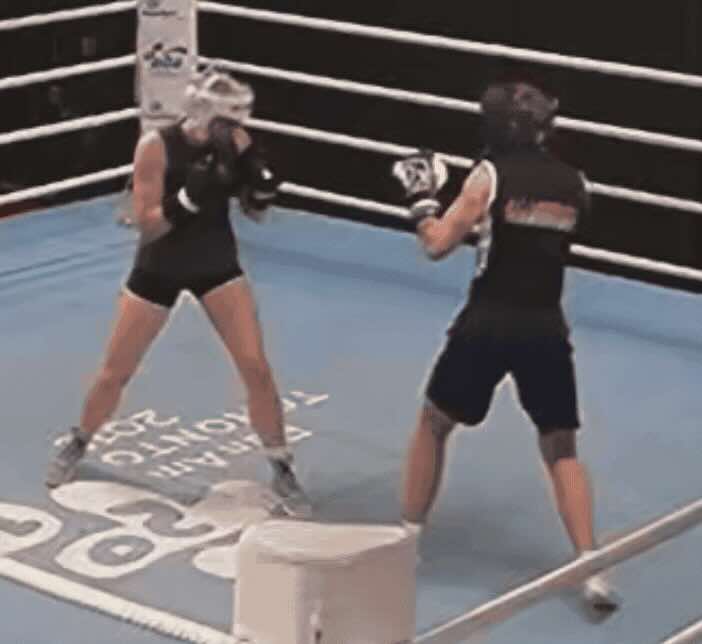}};
\node at (-4,-0.35) {\vdots};
\node[inner sep=0pt] (2) at (-4,-1.5) {\includegraphics[width=.1\textwidth]{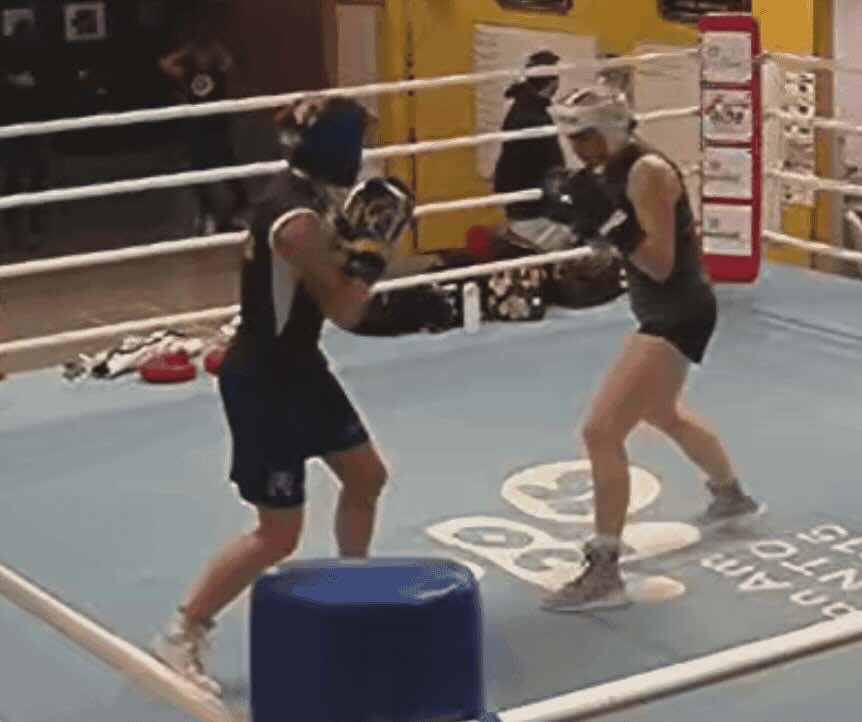}};

\node[process3] (yolo) at (-2.2,-0.4) {Yolov8};
\draw[vecArrow] (-3.1,-0.4) tonode[anchor=west, rotate=-90] {{$\mathrm{view}_j$}} (yolo);
\node[process3] (xmem) at (-1,-0.4) {XMem};
\node[process3] (vitpose) at (1,-0.4) {ViT Pose};
\draw[vecArrow] (yolo) tonode[anchor=west, rotate=-90] {$\mathbf{bboxes}$} (xmem) ;
\node[process3] (epipolar) at (0,-0.4){epipolar constraints};
\draw[vecArrow] (xmem) tonode[anchor=west, rotate=-90] {$\mathbf{ids}$} (epipolar);

\draw[vecArrow] (epipolar) tonode[anchor=west, rotate=-90] {$\mathrm{id}$} (vitpose);

\draw[vecArrow] (vitpose) tonode[anchor=west, rotate=-90] {$\mathbf{J}_{2D}$} (tri);

\node[process] (kinematics) at (8.5,-0.35) {Kinematics\\Optimization};
\node[process2] (prior) at (8.5,0.75) {GMM Prior, Vposer};
\node[process2] (loss) at (8.5,-1.5) {$\mathrm{L}_\text{2D}$, $\mathrm{L}_\text{3D}$, $\mathrm{L}_\text{reg}$, $\mathrm{L}_\text{smooth}$};
\draw[vecArrow] (loss) to (kinematics);
\draw[vecArrow] (prior) to (kinematics);
\draw[vecArrow] (tri) tonode[anchor=south] {$\mathbf{J}_{2D}$, $\mathbf{J}_{3D}$} (kinematics);

\node[inner sep=0pt] (smpl2) at (12,-0.35) {\includegraphics[width=.1\textwidth]{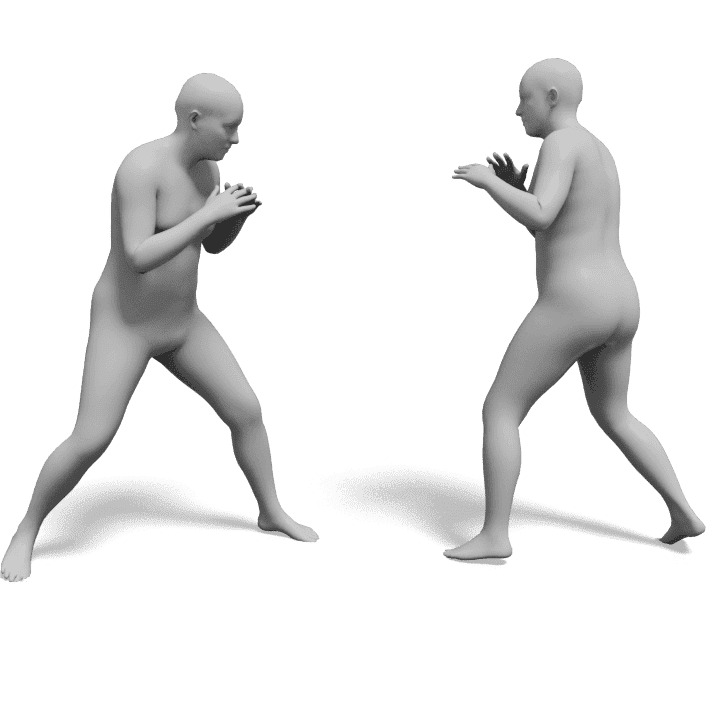}};
\draw[vecArrow] (kinematics) tonode[anchor=south] {\scriptsize{$\mathbf{\theta}$, $\mathbf{\beta}$}} (smpl2);
\node[inner sep=0pt] (smpl) at (-4,-4){\includegraphics[width=.1\textwidth]{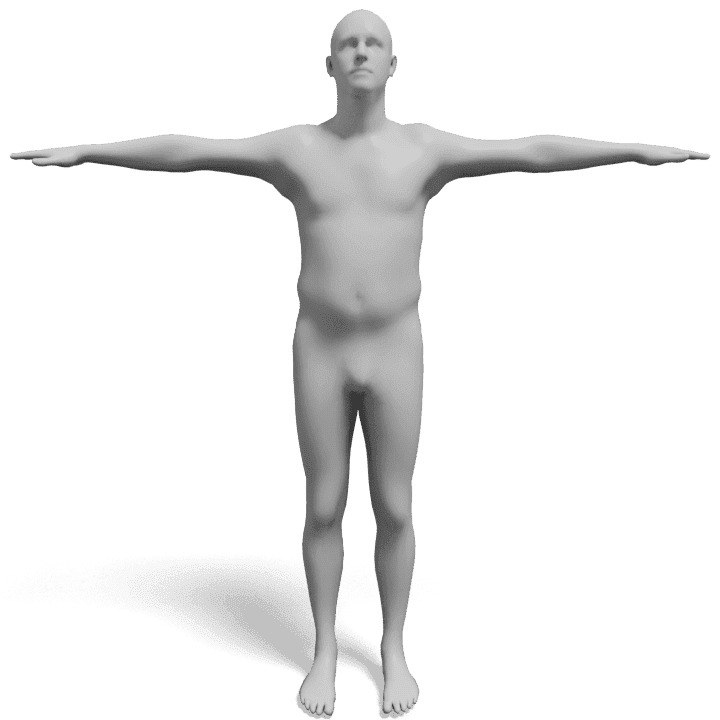}};
\node[decision] (retarget) at (0,-4) {Physics Body Model Generator};
\draw[vecArrow] (smpl) to (retarget);

\node[inner sep=0pt] (humanoid) at (3.5,-4){\includegraphics[width=.1\textwidth]{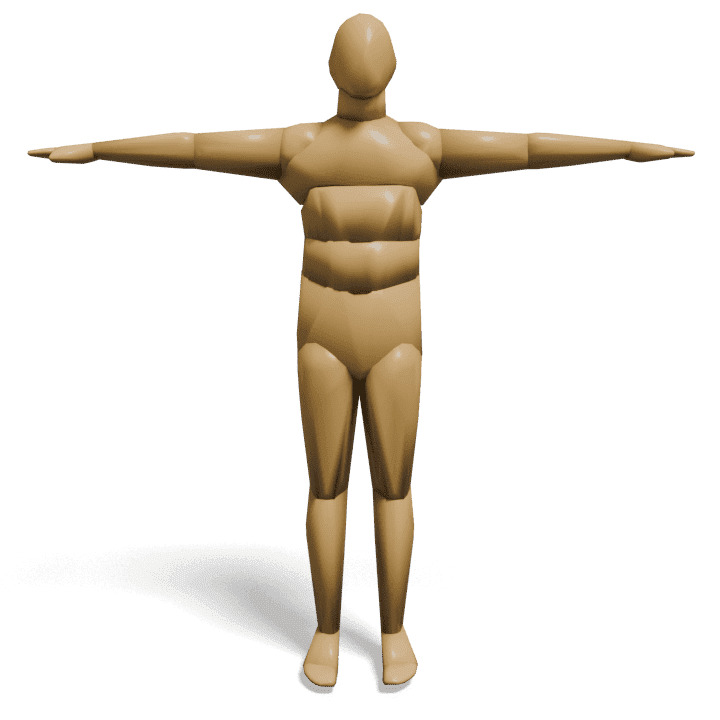}};
\draw[vecArrow] (retarget) to (humanoid);
\node[inner sep=0,minimum size=0] (k) at (6.5, -2.3) {};
\node[process] (lqr) at (7.5,-4) {Multi-Person iLQR Optimization};
\draw[vecArrow] (humanoid) to (lqr);
\draw[vecArrow1] (smpl2) |- (4.1,-2.3);
\draw[vecArrow] (7.5, -2.3) tonode[anchor=west] {$\mathbf{q}, \mathbf{v}, \mathbf{p}$} (lqr);
\node[inner sep=0,minimum size=0] (k1) at (-1, -2.3) {};
\draw[vecArrow1] (4.4, -2.3) to (-0.24, -2.3);
\draw[vecArrow] (0, -2.28) tonode[anchor=west] {\scriptsize{$\mathbf{\beta}$}} (retarget);
\draw[innerWhite] (smpl2) |- (4.22,-2.3);
\draw[innerWhite] (k) -- (lqr);
\draw[innerWhite] (4.4, -2.3) -- (-0.18, -2.3);
\node[inner sep=0pt] (final) at (12,-4){\includegraphics[width=.125\textwidth]{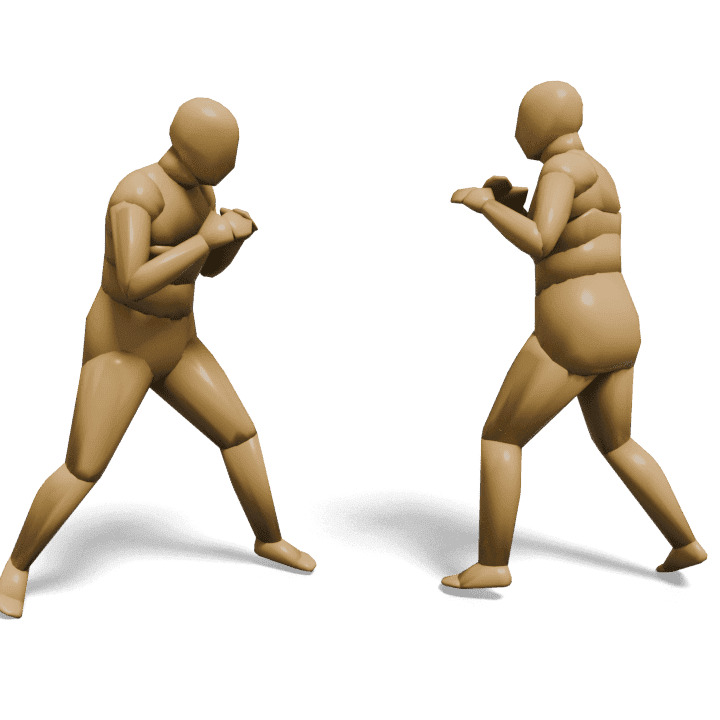}};
\draw[vecArrow] (lqr) to (final);

\end{tikzpicture}
  }
  \caption{Overview of our proposed pipeline. Initially, bounding boxes ($\mathbf{bboxes}$) and robust identity tracking ($\mathrm{id}$) are generated for each individual in the scene. These tracking results facilitate accurate 2D pose estimation ($\mathbf{J}_{2D}$) using ViTPose~\cite{xu2022vitpose}. The subsequent triangulation process yields smoothed 3D keypoints ($\mathbf{J}_{3D}$). These keypoints serve as input to the kinematic optimization stage, which outputs SMPL model parameters ($\mathbf{\theta}$, $\mathbf{\beta}$). Finally, the optimized 3D joint positions ($\mathbf{p}$), pose states ($\mathbf{q}$), and velocity states ($\mathbf{v}$), produced by our custom physics-based humanoid model generator, guide a model predictive controller employing the iLQR optimizer~\cite{howell2022predictivesampling} to eliminate physical artifacts and refine motion quality.}
  \label{fig:3d}
\end{figure*}

Figure~\ref{fig:3d} illustrates the overall architecture of our proposed framework, comprising four key stages: multi-view tracking, weighted triangulation, kinematic optimization, and dynamics-based trajectory refinement. Each stage is detailed in Sections \ref{subsec:multi-view-tracking} through \ref{subsec:dynamics}.

\subsection{Multi-frame Multi-view Tracking IDs}\label{subsec:multi-view-tracking}

Consistent tracking across multiple frames and views is crucial for accurate motion reconstruction, particularly when camera setups are sparse. Limited overlapping fields of view and frequent occlusions pose significant challenges for reliably identifying and tracking the same individual across multiple camera views.

To achieve coherent tracking identities ($\mathbf{ids}$), we utilize XMem~\cite{cheng2022xmem}, a memory-based video segmentation method. This approach generates persistent identity labels within each single view, ensuring temporal consistency. Subsequently, we compute the centroids of the obtained bounding boxes ($\mathbf{bboxes}$) and associate their assigned identities ($\mathbf{ids}$) across multiple views, thus assigning unified identities for each individual in the scene.

To enhance cross-view matching reliability, we leverage epipolar constraints~\cite{chen2020cross, kocabas2019selfsupervised}. These geometric constraints significantly narrow the matching search space by constraining centroid matching along epipolar lines defined by calibrated camera pairs, effectively reducing false associations.

\begin{figure}[H]
    \centering
    
    \scalebox{0.2}{
    
    \begin{overpic}[scale=1.5]{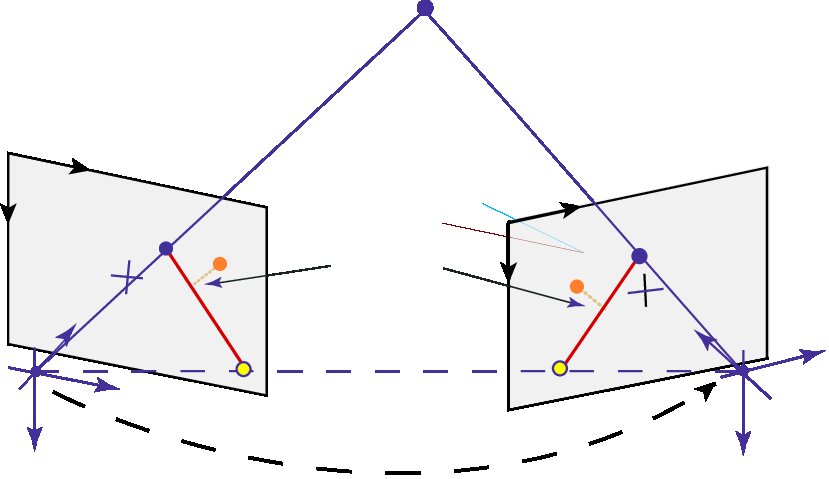}
    \put(15,10){\fontsize{7}{16}\selectfont$\mathrm{X}_1$}
    \put(0,2){\fontsize{7}{16}\selectfont$\mathrm{Y}_1$}
    \put(9,16){\fontsize{7}{16}\selectfont$\mathrm{Z}_1$}
    \put(-2,30){\fontsize{7}{16}\selectfont$\mathrm{U}_1$}
    \put(8,39){\fontsize{7}{16}\selectfont$\mathrm{V}_1$}
    \put(67, 34){\fontsize{7}{16}\selectfont$\mathrm{U}_2$}
    \put(57, 25){\fontsize{7}{16}\selectfont$\mathrm{V}_2$}
    \put(-2,10){\fontsize{7}{16}\selectfont$\mathrm{C}_1$}
    \put(92,11){\fontsize{7}{16}\selectfont$\mathrm{C}_2$}
    \put(99,15){\fontsize{7}{16}\selectfont$\mathrm{X}_2$}
    \put(91,5){\fontsize{7}{16}\selectfont$\mathrm{Y}_2$}
    \put(82,16){\fontsize{7}{16}\selectfont$\mathrm{Z}_2$}

    \put(44,-3){\fontsize{9}{16}\selectfont $\mathrm{R}$}
    \put(17, 18){\fontsize{9}{16}\selectfont$\mathrm{l}_{c_1}(\mathbf{p}_2)$}
    \put(72, 16){\fontsize{9}{16}\selectfont$\mathrm{l}_{c_2}(\mathbf{p}_1)$}

    \put(44,13.5){\fontsize{9}{16}\selectfont $\mathrm{t}$}
    \put(22,27){\fontsize{7}{16}\selectfont$\mathbf{p}_1=(\mathrm{u}_1,\mathrm{v}_1)$}
    \put(63,25){\fontsize{7}{16}\selectfont$\mathbf{p}_2=(\mathrm{u}_2,\mathrm{v}_2)$}
    \put(40,25){\fontsize{10}{16}\selectfont$d_e(\mathbf{p}_1, \mathbf{p}_2)$}
    \put(46,60){\fontsize{10}{16}\selectfont$\mathrm{(x, y, z)}$}
    \end{overpic}}
    \caption{Illustration of the epipolar geometry distance for a pair of cross view 2D positions.}
    \label{fig:epipolar}
\end{figure}

We quantify cross-view consistency through epipolar distances as shown in Figure~\ref{fig:epipolar}. Given the perspective projection matrix for camera $j$ defined as
\[
  \mathbf{P}_j = \mathbf{K}_j \bigl[\, \mathbf{R}_j \,\big|\;\mathbf{t}_j \bigr],
\]
where $\mathbf{K}_j$ represents the intrinsic matrix, $\mathbf{R}_j \in \mathbb{R}^{3\times 3}$ denotes rotation, and $\mathbf{t}_j \in \mathbb{R}^3$ denotes translation, we compute the fundamental matrix relating two camera views as
\[
  \mathbf{F}_{21} = \mathbf{K}_2^{-\mathrm{T}} \, \mathbf{R} \,\mathbf{K}_1^\mathrm{T} \, \bigl[\mathbf{K}_1 \,\mathbf{R} \,\mathbf{t}\bigr]_\times,
\]
where $[\,\cdot\,]_\times$ indicates the skew-symmetric matrix operator.
Given detected centroids $\mathbf{p}_1$ in view 1 and $\mathbf{p}_2$ in view 2, we define the epipolar distance~\cite{wohler20093d} as
\begin{equation}
   d_e(\mathbf{p}_1, \mathbf{p}_2) \;=\; d\bigl(\mathbf{p}_1,l_{\mathrm{c}_1}\bigr) \;+
   \; d\bigl(\mathbf{p}_2,l_{\mathrm{c}_2}\bigr),
\end{equation}
where $l_{\mathrm{c}_1}=\mathbf{F}_{12}\,\mathbf{p}_1$ and $l_{\mathrm{c}_2}=\mathbf{F}_{21}\,\mathbf{p}_2$ represent the epipolar lines. The function $d(\cdot,\cdot)$ calculates the perpendicular distance from a centroid to its corresponding epipolar line. By minimizing $d_e(\mathbf{p}_1,\mathbf{p}_2)$ across all camera pairs, we achieve consistent centroid associations and unified identity labels across multiple views and frames.

\subsection{Weighted Triangulation and Filtering}\label{subsec:triangulation}

We perform weighted triangulation to estimate the 3D positions of keypoints from their corresponding 2D detections across all $N$ cameras. For each joint, we solve the following weighted linear system:
\begin{equation}
\begin{bmatrix}
\mu_{1} (\Proj_{11} - \mathrm{u}_1 \Proj_{31}) \\
\mu_{1} (\Proj_{21} - \mathrm{v}_1 \Proj_{31}) \\
\vdots \\
\mu_{N} (\Proj_{1N} - \mathrm{u}_N \Proj_{3N}) \\
\mu_{N} (\Proj_{2N} - \mathrm{v}_N \Proj_{3N}) \\
\end{bmatrix}
\begin{bmatrix}
\mathrm{x} \\
\mathrm{y} \\
\mathrm{z} \\
1
\end{bmatrix}
=
\begin{bmatrix}
0\\
\vdots \\
0
\end{bmatrix} \,,
\label{eq:tri}
\end{equation} 
where $\Proj_{ij}$ denotes the $i$-th row of the projection matrix $\mathbf{P}_{j}$ for camera $j$, $(\mathrm{u}_j, \mathrm{v}_j)$ represents the detected 2D joint coordinates, and $\mu_j$ is the average detection confidence for camera $j$. Cameras providing higher-confidence detections thus have increased influence in the triangulation process. We solve Eq.\ref{eq:tri} efficiently using Singular Value Decomposition (SVD)\cite{image_processing}.

To handle potential triangulation outliers and missing data, we interpolate and smooth each joint's trajectory using cubic spline interpolation. In scenarios where triangulation fails, we employ an extended Kalman filter (EKF) leveraging the joint's velocity, acceleration, and positional constraints to estimate stable and accurate 3D trajectories. This robust filtering ensures high-quality 3D reconstructions even from sparse and noisy multi-view detections.

\subsection{Kinematics Optimization from Multi-View Keypoints}\label{subsec:kinematics}

Our kinematics optimization leverages multi-view 2D keypoints ($\mathbf{J}_{\text{2D}}$) and triangulated 3D keypoints ($\mathbf{J}_{\text{3D}}$) to fit an SMPL model~\cite{SMPL2015loper}. This fitting process minimizes discrepancies between the SMPL joints and observed keypoints, incorporating temporal smoothness and human motion priors to ensure natural and coherent motion trajectories.

The SMPL shape parameters ($\beta \in \mathbb{R}^{10}$) for each individual are initialized from triangulated 3D keypoints, optimizing to minimize differences between predicted and observed limb lengths. Subsequently, pose parameters ($\mathbf{\theta} \in \mathbb{R}^{72}$) are estimated through optimization.

We formulate this optimization as the minimization of the following objective using the LBFGS optimizer~\cite{NoceWrig06} with history size 100, learning rate 1.0, and strong Wolfe conditions:
\begin{align}
\min_{\theta} \quad & w_1 \, \mathrm{L}_\text{2D} + w_2 \, \mathrm{L}_\text{3D} + w_3 \, \mathrm{L}_\text{reg} + w_4 \, \mathrm{L}_\text{smooth} \nonumber \\
& + w_5 \, \mathrm{L}_{\text{GMM}} + w_6 \, \mathrm{L}_{\text{Vposer}} \,\,.
\label{eq:loss}
\end{align}
with weights $w_1=0.001$, $w_2=1.0$, $w_3=0.01$, $w_4=0.001$, $w_5=0.0001$, $w_6=0.0001$ determined empirically.

The \textbf{2D re-projection loss} $\mathrm{L}_\text{2D}$ aligns projected 3D joints with detected 2D keypoints~\cite{pavlakos2019expressive}:
\begin{equation}
 \mathrm{L}_\text{2D} = \sum\limits_{\text{j} \in \text{$\mathcal{V}$}} \, \sum\limits_{\text{i} \in \text{$\mathbf{J}_{2D}$}} c_\text{j,i} \rho (\mathrm{J}_{\text{proj}_{j,i}} - \mathrm{J}_{2D_{j,i}})  \,,
 \label{eq:loss2d}
\end{equation}
where $c_{j,i}$ denotes confidence scores, we only consider 2D keypoints with confidence higher than 0.7, and $\rho$ is the robust Geman-McClure error function. The projection is given by $\mathrm{J}_{\text{proj}_{j,i}} = \mathbf{P}_j \mathrm{J}_i(\theta, \beta)$, where $\mathbf{P}_j$ is the camera projection matrix, and $\mathbf{P}_j$ and $\mathrm{J}(\theta, \beta) = \mathbf{W} \,\mathcal{M}(\theta, \beta)$ is the joint mapping from mesh vertices $\mathcal{M}$ we only consider 2D keypoints.

The \textbf{3D alignment loss} $\mathrm{L}_\text{3D}$ ensures consistency with triangulated 3D keypoints:
\begin{equation}
  \mathrm{L}_\text{3D} = \sum\limits_{i \in \text{$\mathbf{J}_{3D}$}} c_i \| \mathrm{J}_i(\theta, \beta) - \mathrm{J}_{3D,i}\|^2 \,,
  \label{eq:loss3d}
\end{equation}
where  $\mathrm{J}_{3D,i}$ is the triangulated joint $i$ and $c_i$ is the confidence of the 3D joint $i$ which obtained from averaging the confidences of the 2D joints. 

The \textbf{smoothness loss} $\mathrm{L}_\text{smooth}$ enforces temporal coherence by penalizing abrupt changes between consecutive frames:
\begin{equation}
  \mathrm{L}_{\text{smooth}} = \sum\limits_{t} \| \theta^t - \theta^{t-1} \|^2 + \| \mathcal{M}(\theta^t, \beta) - \mathcal{M}(\theta^{t-1}, \beta) \|^2 \,.
  \label{eq:loss-smooth} 
\end{equation} 

The regularization terms $\mathrm{L}_\text{reg}$, $\mathrm{L}_{\text{GMM}}$, and $\mathrm{L}_{\text{Vposer}}$ further guide the optimization toward plausible human poses by penalizing unnatural or exaggerated configurations, significantly reducing jitter and noise. 

Finally, the \textbf{prior loss} term penalizes unnatural poses of the SMPL skeleton. Specifically, we incorporate two priors into the loss function: the Gaussian Mixture Model (GMM) prior~\cite{bogo2016keep} and the Vposer prior~\cite{pavlakos2019expressive}. These priors guide the optimization towards realistic human poses and reduce jittering artifacts. The losses are defined as:
\[
\mathrm{L}_{\text{GMM}} = \frac{1}{N} \sum_{i=1}^{N} \text{GMM}(\theta_i, \beta), \quad \mathrm{L}_{\text{Vposer}} = \frac{1}{N} \sum_{i=1}^{N} (\text{z}(\theta_i)^{2}) \,,
\]
where the function $\text{z}(\cdot)$ maps the full pose into a latent space defined by the Vposer prior.

\subsection{Multi-person Dynamic Optimization}\label{subsec:dynamics}

Although the kinematic optimization stage can estimate plausible full-body poses for multiple individuals, artifacts such as foot sliding, inter-body penetrations, and ground penetration can frequently occur. To address these physical implausibilities, we introduce a multi-person dynamics optimization stage. This stage employs individualized physics-based humanoid models, parameterized by the shape parameters ($\beta$), significantly improving motion realism and robustness.

\noindent\textbf{Physics-based Humanoid.} To ensure the physics humanoid accurately reflects the joint positions estimated by the SMPL model, we implement a custom physics-based humanoid generator. This generator utilizes joint positions and SMPL mesh landmarks at a T-pose to create articulated rigid-body models with convex mesh-shaped collision geometries. Convex hull mesh dimensions are derived directly from SMPL mesh segmentation.

Our physics humanoid model contains 69 joint-angle degrees of freedom (DOFs) alongside a 6-DOF free root joint. The joints are primarily modeled as hinge joints, allowing realistic biomechanical articulation. The base skeleton of the humanoid follows the SMPL skeleton with an assigned average human body density of $985 , \text{kg/m}^3$.

\begin{figure}[tb]
    \centering
    
    \begin{minipage}[b]{0.24\columnwidth}
        \includegraphics[width=\linewidth]{figures/smpl-tall.jpg}
    \end{minipage}%
    \begin{minipage}[b]{0.24\columnwidth}
        \includegraphics[width=\linewidth]{figures/humanoid-tall.jpg}
    \end{minipage}%
    \begin{minipage}[b]{0.24\columnwidth}
        \includegraphics[width=\linewidth]{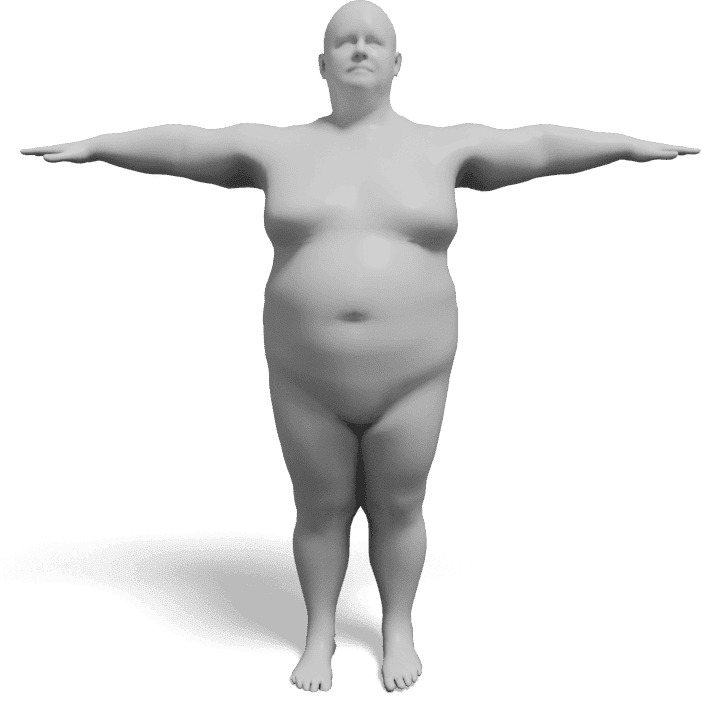}
    \end{minipage}%
    \begin{minipage}[b]{0.24\columnwidth}
        \includegraphics[width=\linewidth]{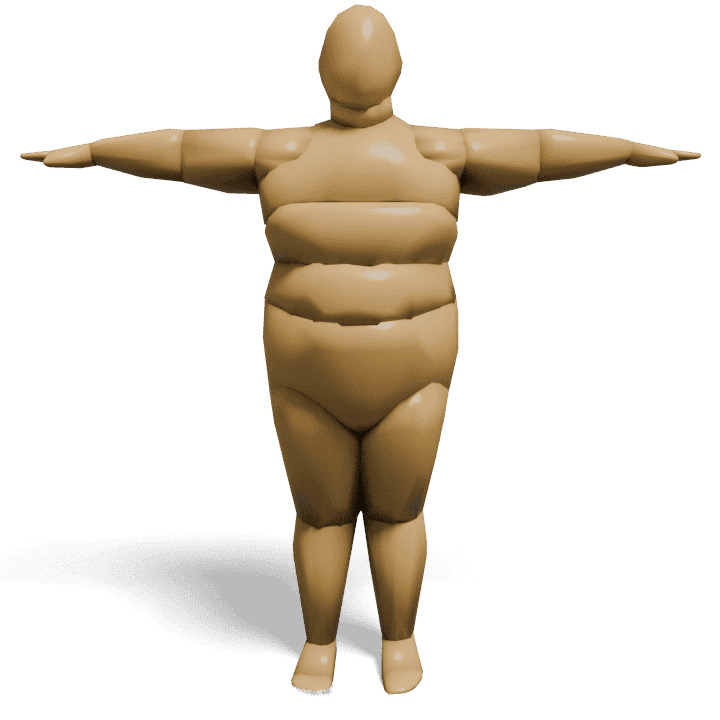}
    \end{minipage}%
    
    \caption{Illustration of SMPL to physics-based humanoid conversion for varying body shapes.}
    \label{fig:humanoid}
\end{figure}

Several examples of physics-based humanoid models generated using this approach are shown in Figure~\ref{fig:humanoid}.

\textbf{Multi-humanoid Model.} We define the combined state of $K$ physics-based humanoids at each timestep $t$ as $\mathbf{x}_t = (\mathbf{q}_t, \mathbf{v}_t)$, where $\mathbf{q}_t \in \mathbb{R}^{63K}$ and $\mathbf{v}_t \in \mathbb{R}^{62K}$ represent the joint rotations and velocities, respectively. While $\mathbf{q}_t$ uses minimal coordinate representation, we compute the 3D joint positions relative to each humanoid's body reference frame, represented by $\mathbf{p}_t \in \mathbb{R}^{21K}$, to guide optimization. The root orientation is encoded as a quaternion. Additional details about the humanoid model are provided in the supplementary material.

\textbf{Dynamical Simulation.} Joint torques actuate our physics-based humanoid models, computed via a physics-informed version of the previous kinematic optimization. Unlike purely kinematic methods, our approach integrates biomechanical constraints and realistic contact interactions using an advanced physics simulator~\cite{todorov2012mujoco}. The humanoid's state evolves according to:
\[
    \mathbf{x}_{t+1} \gets f(\mathbf{x}_t, \mathbf{u}_t),
\]
where $\mathbf{u}_t \in \mathbb{R}^{56K}$ represents joint torques excluding direct root actuation.

\textbf{Trajectory Optimization.}
Our trajectory optimization leverages iterative Linear Quadratic Regulation (iLQR)~\cite{howell2022predictivesampling}, minimizing the objective:
\begin{equation}
 \min_{\mathbf{u}_{0:T}} \sum_{t=0}^{T} w_1 \mathrm{L_{\text{reg},t}} + w_2 \mathrm{L}_{\text{p},t} + w_3 \mathrm{L}_{\text{v},t} + w_4 \mathrm{L}_{\text{collision},t},
 \label{eq:mpc_opt}
\end{equation}
with empirically determined weights $w_1=0.001$, $w_2=10$, $w_3=0.1$, $w_4=20$. The terms $\mathrm{L}_{\text{reg}}$ includes $\|\mathbf{v}_t\|^2$ and $\|\mathbf{u}_t\|^2$ for minimizing velocity and control values which ensure smooth, realistic motion by penalizing deviations from reference positions and velocities, and excessive joint torques and velocities. The optimized trajectory is refined iteratively via feedback:
\[
\Delta\mathbf{u}_t = \mathbf{K}_t\Delta\mathbf{x}_t + \alpha \mathbf{k}_t,
\]
where the feedback gain $\mathbf{K}_t$ and offset $\mathbf{k}_t$ are updated at each iteration to achieve convergence. We employ a parallel line search for optimal step size $\alpha$ within $[\alpha_{\text{min}}, 1]$. Further details are provided in Tassa et al.~\cite{tassa2012trajectoryoptimization}.
\\

To prevent persistent inter-penetrations from kinematic estimates, we add a collision penalty term in our trajectory optimization: 
\[
\mathrm{L}_{\text{collision},t} = \sum_{(i,j)\in \mathcal{C}_t} \phi\bigl(d_{\mathrm{overlap}}(i,j)\bigr),
\]
where \(\mathcal{C}_t\) denotes colliding parts at time \(t\) with \(d_{\mathrm{overlap}}(i,j)\) quantifying penetration depth and \(\phi\) growing rapidly (e.g., quadratically) with overlap; additionally, the physics engine enforces Coulomb friction and non-penetration via contact manifolds, ensuring physically compliant responses.

\section{Experiments and Results}
\label{sec:results}

\begin{figure*}[htbp]
    \centering
    \scriptsize
    \begin{subfigure}[t]{0.32\linewidth}
        \centering
        \textbf{Campus Dataset} \\[0.2em]
        \includegraphics[width=\linewidth]{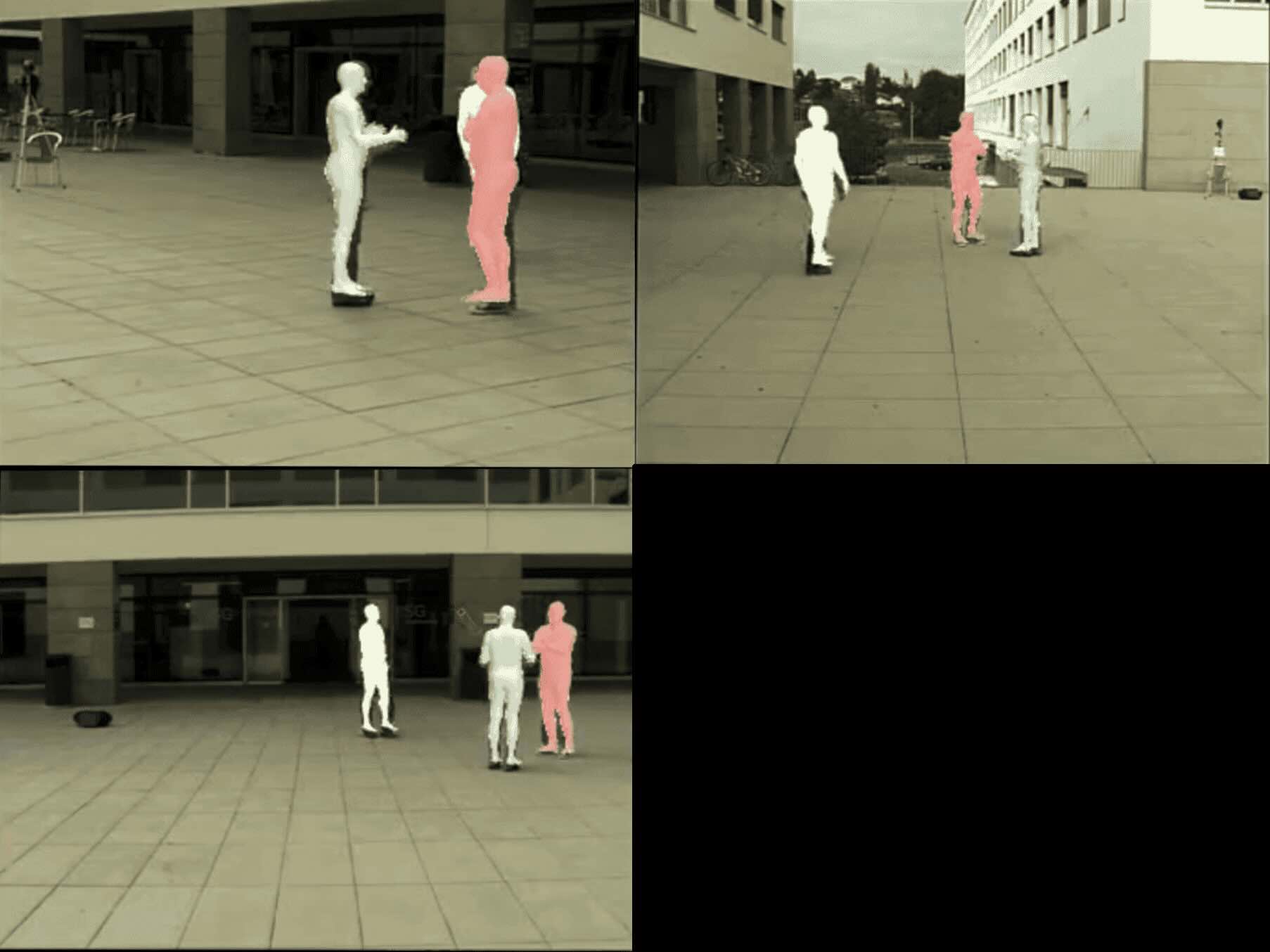}
    \end{subfigure}
    \hspace{0.01\textwidth}
    \begin{subfigure}[t]{0.32\linewidth}
        \centering
        \textbf{Shelf Dataset} \\[0.2em]
        \includegraphics[width=\linewidth]{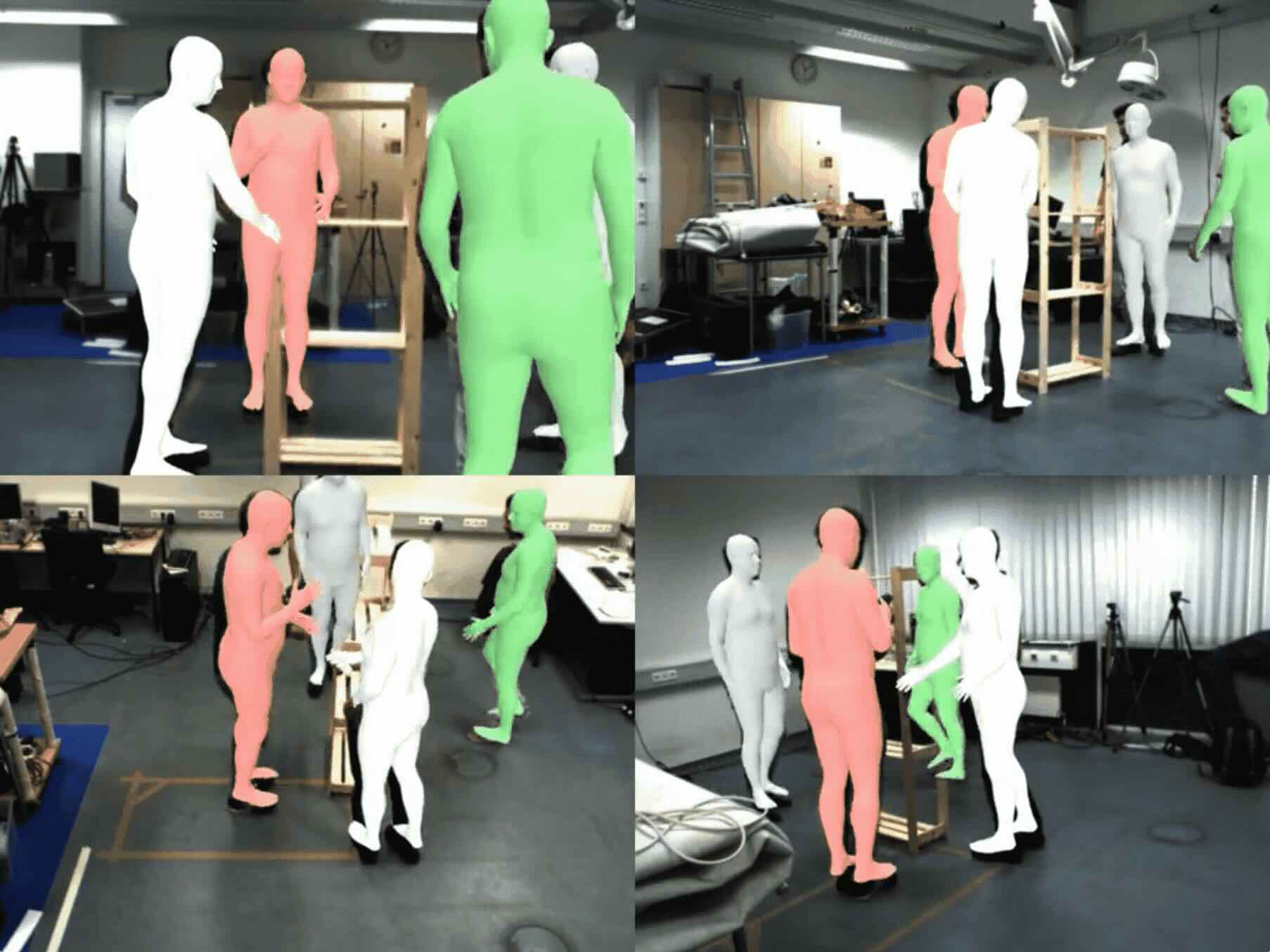}
    \end{subfigure}
    \hspace{0.01\textwidth}
    \begin{subfigure}[t]{0.32\linewidth}
        \centering
        \textbf{Supplementary Dataset} \\[0.2em]
        \includegraphics[width=\linewidth]{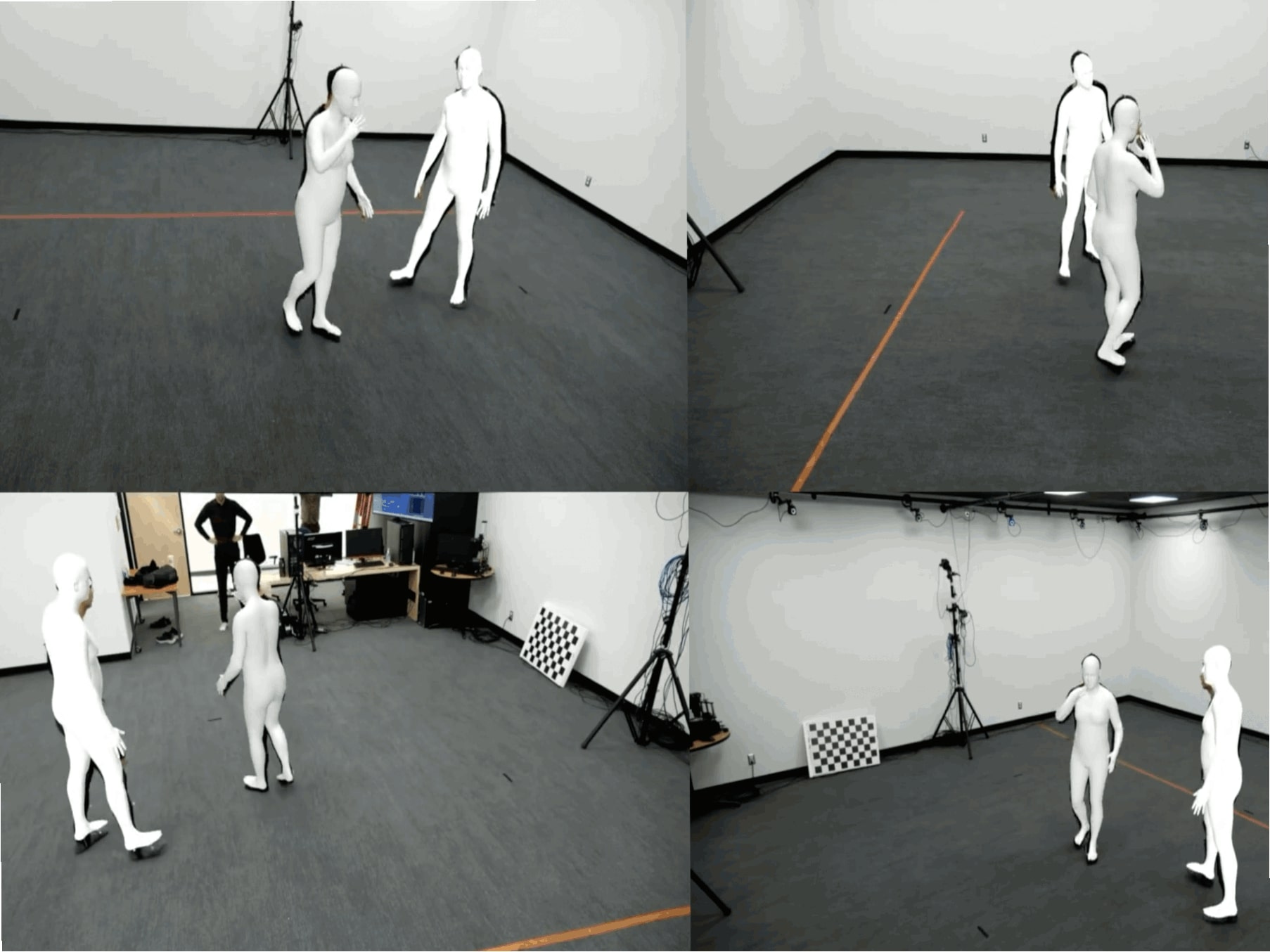}
    \end{subfigure}
    \caption{\textbf{Qualitative results} on various datasets. Our method accurately reconstructs human motion and outperforms or matches state-of-the-art kinematic methods.}
    \label{fig:qualitative}
\end{figure*}

In this section, we report results from a series of experiments on two established multi-view benchmarks, \textit{Campus}~\cite{fleuret2007multicamera} and \textit{Shelf}~\cite{belagiannis2016pami3dpictorial}, as summarized in Table~\ref{tab:benchmark}. We also evaluate our approach on two monocular datasets, \textit{CHI3D}~\cite{fieraru2020threedHumanInteractions} and \textit{Hi4D}~\cite{yin2023hi4d}, reported in Table~\ref{tab:sota}. Three ablation studies demonstrate the role of physics constraints in our final results. Figure~\ref{fig:qualitative} shows example frames from various datasets. 

\paragraph{Supplementary Material.}
Our supplementary video and additional documentation offer deeper insights into our approach. Notably, we present video clips of two boxers in practice, exemplifying the robustness of our pipeline in complex scenarios involving close interactions. We also include an ablation study on each subsystem of our pipeline, validating its design.

\paragraph{Multi-view Benchmarks.}
We first compare our method to state-of-the-art multi-view approaches on the \textit{Campus}~\cite{fleuret2007multicamera} (three cameras, three subjects) and \textit{Shelf}~\cite{belagiannis2016pami3dpictorial} (five cameras, four subjects) datasets. Table~\ref{tab:benchmark} lists Percentage of Correctly estimated body Parts (PCP)~\cite{fleuret2007multicamera}, showing that on the Shelf dataset, our approach achieves the highest average PCP. On Campus, we are competitive with the best existing methods. For a detailed explanation of the metrics and additional ablations, see the supplementary material. As no dedicated multi-view physics-based approach is available for multi-person settings, we compare against existing multi-view and some monocular multi-person physics-based methods.

\paragraph{Multi-Person Interaction Datasets.}
To evaluate our method in comparison with the existing multi-person physics-based trackers, we apply our dynamics optimization using monocular footage. Keypoints and pose estimates are extracted using SLAHMR~\cite{ye2023decoupling}, which is employed in Multiphys~\cite{Ugrinovic2024MultiPhys}. We utilized the same evaluation datasets as in ~\cite{Ugrinovic2024MultiPhys}. These datasets exhibit significant inter-person interactions, where methods based solely on kinematics often struggle. We evaluated our method using two datasets with varying levels of interaction.

The CHI3D dataset~\cite{fieraru2020threedHumanInteractions} also includes some close interactions, and includes 127 motion sequences of five pairs of subjects performing everyday actions such as posing, pushing, and hugging. Each sequence is annotated with action labels and ground-truth 3D poses in the SMPL format, using four camera views. The Hi4D data set~\cite{yin2023hi4d} involves more intense interactions, with 100 short motion sequences that contain close interactions and high contact ratios between subjects performing actions such as hugging, posing, dancing and playing sports. These sequences were captured with up to eight cameras and involve 20 unique pairs of participants.
As shown in Table~\ref{tab:sota}, our method outperforms existing methods in most physics-based metrics, except for penetration.
\begin{table}[htbp]
\centering
\caption{Quantitative comparison on our supplementary dataset.}
\label{tab:physics}
\resizebox{0.9\columnwidth}{!}{%
  \begin{tabular}{lcccc}
    \toprule
    Method    & $e_{\text{MPJPE}} \downarrow$ & $e_{\text{foot},z}\downarrow$ & $e_{\text{foot},v_{xy}}\downarrow$ & $e_{\text{smooth}}\downarrow$ \\
    \midrule
    Kinematics & 41.2 & 16.4 & 2.2 & 6.1 \\
    Dynamics   & 38.4 & 8.1  & 0.3 & 4.6 \\
    \bottomrule
  \end{tabular}%
}
\end{table}
\begin{table*}[htb!]
  \centering
  %
  \begin{minipage}{0.48\linewidth}
    \centering
    \caption{Comparison of PCP [\% $\uparrow$] on the supplementary dataset.}
    \label{tab:supp}
    \resizebox{\linewidth}{!}{%
      \begin{tabular}{@{}r|cccc|cccc|cccc@{}}
        \toprule
        Method & \multicolumn{4}{c|}{4 cameras} & \multicolumn{4}{c|}{3 cameras} & \multicolumn{4}{c}{2 cameras} \\
        \cmidrule(r){2-5} \cmidrule(lr){6-9} \cmidrule(l){10-13}
        & Actor 1 & Actor 2 & Avg. & & Actor 1 & Actor 2 & Avg. & & Actor 1 & Actor 2 & Avg. & \\
        \midrule
        Triangulation & 98.4 & 97.6 & 98.0 & & 98.0 & 97.1 & 97.5 & & 93.2 & 84.1 & 88.6 & \\
        Kinematics    & 99.2 & 98.9 & 99.0 & & 98.2 & 97.6 & 97.9 & & 93.6 & 84.4 & 89.0 & \\
        Dynamics      & 98.6 & 98.2 & 98.4 & & 97.9 & 94.7 & 96.3 & & 91.2 & 89.9 & 90.5 & \\
        \bottomrule
      \end{tabular}
    }
  \end{minipage}
  \hfill
  %
  \begin{minipage}{0.48\linewidth}
    \centering
    \caption{Comparing single-person vs. multi-person dynamics optimization on CHI3D~\cite{fieraru2020threedHumanInteractions}.}
    \label{tab:ablation1}
    \resizebox{\linewidth}{!}{%
      \begin{tabular}{lcccc}
        \toprule
        Method & Pen. & WA-MPJPE & W-MPJPE & PA-MPJPE (joint) \\ 
        \midrule 
        Single-person Physics & 114.9 & 117.3 & 174.5 & 92.1 \\
        Multi-person Physics  & 18.7  & 86.4  & 156.4 & 75.2 \\
        \bottomrule
      \end{tabular}
    }
  \end{minipage}
\end{table*}

We also ablate the effect of single-person vs. multi-person physics optimization on CHI3D in Table~\ref{tab:ablation1}. A single-person model cannot fully resolve interpersonal penetrations and yields larger pose errors. In contrast, our multi-person optimization significantly reduces these artifacts and improves accuracy. Detailed definitions of physics metrics appear in the supplementary material.

\begin{table}[htbp]
  \centering
  \begin{minipage}{\columnwidth}
    \caption{Comparison of PCP~[\% $\uparrow$] on the Campus and Shelf datasets.}
    \label{tab:benchmark}
    \resizebox{0.9\columnwidth}{!}{%
      \begin{tabular}{@{}r|cccc|cccc@{}}
        \toprule
        Method & \multicolumn{4}{c|}{Campus} & \multicolumn{4}{c}{Shelf} \\
        \cmidrule(r){2-5} \cmidrule(l){6-9}
               & Actor 1 & Actor 2 & Actor 3 & Avg. & Actor 1 & Actor 2 & Actor 3 & Avg.\\
        \midrule
        \citet{Belagiannis_2014_CVPR}        & 82.0  & 72.4  & 73.7  & 75.8 & 66.1  & 65.0  & 83.2  & 71.4 \\
        \citet{belagiannis2016pami3dpictorial}         & 93.5  & 75.7  & 84.4  & 84.5 & 75.3  & 69.7  & 87.6  & 77.5 \\
        \citet{ershadi2018multiple}                     & 94.2  & 92.9  & 84.6  & 90.6 & 93.3  & 75.9  & 94.8  & 88.0 \\
        \citet{dong2019mvpose}                          & 97.6  & 93.3  & 98.0  & 96.3 & 98.8  & 94.1  & 97.8  & 96.9 \\
        \citet{Huang2020endtoenddynamicmatching}       & 98.0  & \textbf{94.8} & 97.4 & 96.7 & 98.8  & 96.2  & 97.2  & 97.4 \\
        \citet{tu2020voxelpose}                         & 97.6  & 93.8  & 98.8  & 96.7 & 99.3  & 94.1  & 97.6  & 97.0 \\
        \citet{lin2021multiviewplanesweep}              & 98.4  & 93.7  & \textbf{99.0} & 97.0 & 99.3  & 96.5  & 98.0  & 97.9 \\
        \citet{wang2021mvp}                 & 99.3  & 95.1  & 97.8  & 97.4 & 98.2  & 94.1  & 97.4  & 96.6 \\ 
        \citet{yang2023unifiedMultiView}                 & 98.2  & 94.6  & 98.2  & 97.0 & 99.5  & 96.0  & 97.7  & 97.7 \\       
        ours (Triangulation)                             & \textbf{99.6} & 92.2 & 97.6  & 96.5 & \textbf{99.8} & 95.4  & \textbf{98.6} & \textbf{97.9} \\ 
        ours (Kinematics)                                & 98.5  & 93.5  & 94.4  & 95.5 & \textbf{99.8} & \textbf{97.6} & \textbf{98.6} & \textbf{98.6} \\
        ours (Dynamics)                                  & 98.1  & 93.6  & 94.2  & 95.3 & 98.9  & \textbf{97.9} & 98.2  & 98.3 \\
        \bottomrule
      \end{tabular}%
    }
  \end{minipage}
\end{table}
\begin{table}[htbp]
  \centering
  \begin{minipage}{\columnwidth}
    \caption{\textbf{Close interaction datasets.} Comparison on CHI3D~\cite{fieraru2020threedHumanInteractions} and Hi4D~\cite{yin2023hi4d} with existing state-of-the-art trackers. Metrics: Penetration (Pen.), Ground Penetration (Gnd Pen.), Skating, Acceleration Error (Acc. Error), WA-MPJPE, W-MPJPE, and PA-MPJPE (joint) in mm.}
    \label{tab:sota}
    \resizebox{0.9\columnwidth}{!}{%
      \begin{tabular}{@{}clccccccc@{}}
        \toprule
                 & Method & Pen. & Gnd Pen. & Skating & Acc. Error & WA-MPJPE & W-MPJPE & PA-MPJPE (joint) \\ 
        \midrule
        \multirow{4}{*}{CHI3D} 
                 & SLAHMR~\cite{ye2023decoupling}         & 139.3 & 4.4  & \textbf{1.0}  & \textbf{6.5}  & 100.9 & 177.1 & 83.5 \\
                 & EmbPose-MP~\cite{Luo2022EmbodiedSceneAware} & 40.2  & 2.6  & 2.8           & 7.7           & 126.7 & 214.7 & 96.5 \\
                 & MultiPhys~\cite{Ugrinovic2024MultiPhys}    & \textbf{18.7} & 3.2  & 2.7        & 7.4           & 98.1  & 174.7 & 80.4 \\
                 & Ours (Monocular)                          & 21.4  & \textbf{1.8} & 2.3      & 6.7           & \textbf{86.4}  & \textbf{156.4} & \textbf{75.2} \\ 
        \midrule
        \multirow{4}{*}{Hi4D} 
                 & SLAHMR~\cite{ye2023decoupling}         & 367.3 & 12.2 & 4.9           & \textbf{6.9}  & \textbf{80.9} & 121.6 & \textbf{69.1} \\
                 & EmbPose-MP~\cite{Luo2022EmbodiedSceneAware} & \textbf{39.8} & 3.8  & \textbf{1.3}  & 12.7          & 115.3 & 148.8 & 92.9 \\
                 & MultiPhys~\cite{Ugrinovic2024MultiPhys}    & 51.1  & 2.4  & 3.5           & 9.6           & 84.8  & \textbf{118.1} & 71.2 \\
                 & Ours (Monocular)                          & 67.2  & \textbf{1.6} & 1.9       & 8.3           & 91.1  & 124.1 & 82.1 \\ 
        \bottomrule
      \end{tabular}
    }
  \end{minipage}
\end{table}

\paragraph{Supplementary Dataset.}
We also collected a custom two-person dataset with close interactions to evaluate foot-ground contacts and overall motion quality. Ground-truth trajectories come from a 24-camera OptiTrack system, and we capture RGB images from up to 4 cameras. The cameras were calibrated with OpenCV~\cite{OpenCVcalib} and then aligned to the motion capture frame. Figure~\ref{fig:qualitative} shows some frames from this dataset.

Table~\ref{tab:supp} demonstrates our system’s robustness across different camera counts, retaining high PCP under fewer viewpoints. Although the final dynamics optimization can slightly reduce PCP due to the rigid-body model, Table~\ref{tab:physics} indicates that physics-based constraints improve the final motion quality by lowering foot sliding and jitter.

\paragraph{Conclusion.}
We introduced a new framework for creating multi-person physics-based animations from multi-view RGB video. Our pipeline comprises robust multi-person tracking, triangulation, kinematics optimization, and a multi-person dynamics optimization. On multi-view benchmarks like Shelf~\cite{belagiannis2016pami3dpictorial}, we achieve state-of-the-art performance, and our approach easily adapts to challenging real-world settings with minimal setup. By combining multi-view geometry with a physics layer, we resolve foot skating, body interpenetrations, and ground collisions. In the future, we will explore more efficient dynamics optimization strategies to further reduce runtime, aiming toward real-time deployment of the entire pipeline.

{
    \twocolumn
    \bibliographystyle{ieeenat_fullname}
    \bibliography{main}

\begin{thebibliography}{64}
\providecommand{\natexlab}[1]{#1}
\providecommand{\url}[1]{\texttt{#1}}
\expandafter\ifx\csname urlstyle\endcsname\relax
  \providecommand{\doi}[1]{doi: #1}\else
  \providecommand{\doi}{doi: \begingroup \urlstyle{rm}\Url}\fi

\bibitem[AlShami et~al.(2023)AlShami, Boult, and
  Kalita]{alshami2023pose2trajectory}
Ali AlShami, Terrance Boult, and Jugal Kalita.
\newblock Pose2trajectory: Using transformers on body pose to predict tennis
  player’s trajectory.
\newblock \emph{Journal of Visual Communication and Image Representation},
  97:\penalty0 103954, 2023.

\bibitem[Andrews et~al.(2016)Andrews, Huerta, Komura, Sigal, and
  Mitchell]{andrews2016real}
Sheldon Andrews, Ivan Huerta, Taku Komura, Leonid Sigal, and Kenny Mitchell.
\newblock Real-time physics-based motion capture with sparse sensors.
\newblock In \emph{Proceedings of the 13th European Conference on Visual Media
  Production (CVMP 2016)}, New York, NY, USA, 2016. Association for Computing
  Machinery.

\bibitem[Belagiannis et~al.(2014)Belagiannis, Amin, Andriluka, Schiele, Navab,
  and Ilic]{Belagiannis_2014_CVPR}
Vasileios Belagiannis, Sikandar Amin, Mykhaylo Andriluka, Bernt Schiele, Nassir
  Navab, and Slobodan Ilic.
\newblock 3d pictorial structures for multiple human pose estimation.
\newblock In \emph{2014 IEEE Conference on Computer Vision and Pattern
  Recognition}, pages 1669--1676, 2014.

\bibitem[Belagiannis et~al.(2016)Belagiannis, Amin, Andriluka, Schiele, Navab,
  and Ilic]{belagiannis2016pami3dpictorial}
Vasileios Belagiannis, Sikandar Amin, Mykhaylo Andriluka, Bernt Schiele, Nassir
  Navab, and Slobodan Ilic.
\newblock 3d pictorial structures revisited: Multiple human pose estimation.
\newblock \emph{IEEE Transactions on Pattern Analysis and Machine
  Intelligence}, 38\penalty0 (10):\penalty0 1929--1942, 2016.

\bibitem[Bogo et~al.(2016)Bogo, Kanazawa, Lassner, Gehler, Romero, and
  Black]{bogo2016keep}
Federica Bogo, Angjoo Kanazawa, Christoph Lassner, Peter Gehler, Javier Romero,
  and Michael~J. Black.
\newblock Keep it smpl: Automatic estimation of 3d human pose and shape from a
  single image.
\newblock In \emph{Computer Vision -- ECCV 2016}, pages 561--578, Cham, 2016.
  Springer International Publishing.

\bibitem[Bridgeman et~al.(2019)Bridgeman, Volino, Guillemaut, and
  Hilton]{bridgeman2019multi}
Lewis Bridgeman, Marco Volino, Jean-Yves Guillemaut, and Adrian Hilton.
\newblock Multi-person 3d pose estimation and tracking in sports.
\newblock In \emph{2019 IEEE/CVF Conference on Computer Vision and Pattern
  Recognition Workshops (CVPRW)}, pages 2487--2496, 2019.

\bibitem[Chang et~al.(2024)Chang, Chang, Shih, Lin, and
  Shih]{chang2024basketball}
Hao-Hsiang Chang, Yu-Hua Chang, Yi-Lung Shih, Cheng-Hsun Lin, and Huang-Chia
  Shih.
\newblock Basketball player action recognition and tracking using r (2+ 1) d
  cnn with spatial-temporal features.
\newblock In \emph{2024 IEEE 13th Global Conference on Consumer Electronics
  (GCCE)}, pages 388--389. IEEE, 2024.

\bibitem[Chen et~al.(2020{\natexlab{a}})Chen, Guo, Li, Lee, and
  Chirikjian]{chen2020multi}
He Chen, Pengfei Guo, Pengfei Li, Gim~Hee Lee, and Gregory Chirikjian.
\newblock Multi-person 3d pose estimation in crowded scenes based on multi-view
  geometry.
\newblock In \emph{Computer Vision – ECCV 2020: 16th European Conference,
  Glasgow, UK, August 23–28, 2020, Proceedings, Part III}, page 541–557,
  Berlin, Heidelberg, 2020{\natexlab{a}}. Springer-Verlag.

\bibitem[Chen et~al.(2020{\natexlab{b}})Chen, Ai, Chen, Zhuang, and
  Liu]{chen2020cross}
Long Chen, Haizhou Ai, Rui Chen, Zijie Zhuang, and Shuang Liu.
\newblock Cross-view tracking for multi-human 3d pose estimation at over 100
  fps.
\newblock In \emph{2020 IEEE/CVF Conference on Computer Vision and Pattern
  Recognition (CVPR)}, pages 3276--3285, 2020{\natexlab{b}}.

\bibitem[Cheng and Schwing(2022)]{cheng2022xmem}
Ho~Kei Cheng and Alexander~G. Schwing.
\newblock Xmem: Long-term video object segmentation with an atkinson-shiffrin
  memory model.
\newblock In \emph{Computer Vision -- ECCV 2022}, pages 640--658, Cham, 2022.
  Springer Nature Switzerland.

\bibitem[Cheng et~al.(2021)Cheng, Wang, Yang, and Tan]{cheng2021topdown}
Yu Cheng, Bo Wang, Bo Yang, and Robby~T. Tan.
\newblock Monocular 3d multi-person pose estimation by integrating top-down and
  bottom-up networks.
\newblock In \emph{CVPR}, 2021.

\bibitem[Distante and Distante(2020)]{image_processing}
Arcangelo Distante and Cosimo Distante.
\newblock \emph{Handbook of Image Processing and Computer Vision: Volume 3:
  From Pattern to Object}.
\newblock Springer, 2020.

\bibitem[Dong et~al.(2022)Dong, Fang, Jiang, Yang, Huang, Bao, and
  Zhou]{dong2019mvpose}
Junting Dong, Qi Fang, Wen Jiang, Yurou Yang, Qixing Huang, Hujun Bao, and
  Xiaowei Zhou.
\newblock Fast and robust multi-person 3d pose estimation and tracking from
  multiple views.
\newblock pages 6981--6992, 2022.

\bibitem[Ershadi-Nasab et~al.(2018)Ershadi-Nasab, Noury, Kasaei, and
  Sanaei]{ershadi2018multiple}
Sara Ershadi-Nasab, Erfan Noury, Shohreh Kasaei, and Esmaeil Sanaei.
\newblock Multiple human 3d pose estimation from multiview images.
\newblock \emph{Multimedia Tools Appl.}, 77\penalty0 (12):\penalty0
  15573–15601, 2018.

\bibitem[Fieraru et~al.(2020)Fieraru, Zanfir, Oneata, Popa, Olaru, and
  Sminchisescu]{fieraru2020threedHumanInteractions}
Mihai Fieraru, Mihai Zanfir, Elisabeta Oneata, Alin-Ionut Popa, Vlad Olaru, and
  Cristian Sminchisescu.
\newblock Three-dimensional reconstruction of human interactions.
\newblock In \emph{2020 IEEE/CVF Conference on Computer Vision and Pattern
  Recognition (CVPR)}, pages 7212--7221, 2020.

\bibitem[Fieraru et~al.(2021)Fieraru, Zanfir, Szente, Bazavan, Olaru, and
  Sminchisescu]{Fieraru2021REMIPS}
Mihai Fieraru, Mihai Zanfir, Teodor~Alexandru Szente, Eduard~Gabriel Bazavan,
  Vlad Olaru, and Cristian Sminchisescu.
\newblock Remips: physically consistent 3d reconstruction of multiple
  interacting people under weak supervision.
\newblock In \emph{Proceedings of the 35th International Conference on Neural
  Information Processing Systems}, Red Hook, NY, USA, 2021. Curran Associates
  Inc.

\bibitem[Fleuret et~al.(2008)Fleuret, Berclaz, Lengagne, and
  Fua]{fleuret2007multicamera}
Francois Fleuret, Jerome Berclaz, Richard Lengagne, and Pascal Fua.
\newblock Multicamera people tracking with a probabilistic occupancy map.
\newblock \emph{IEEE Transactions on Pattern Analysis and Machine
  Intelligence}, 30\penalty0 (2):\penalty0 267--282, 2008.

\bibitem[Gong et~al.(2023)Gong, Foo, Fan, Ke, Rahmani, and
  Liu]{gong2023diffpose}
Jia Gong, Lin~Geng Foo, Zhipeng Fan, Qiuhong Ke, Hossein Rahmani, and Jun Liu.
\newblock Diffpose: Toward more reliable 3d pose estimation.
\newblock In \emph{CVPR}, 2023.

\bibitem[Gärtner et~al.(2022{\natexlab{a}})Gärtner, Andriluka, Coumans, and
  Sminchisescu]{Gartner2022Differentiable}
Erik Gärtner, Mykhaylo Andriluka, Erwin Coumans, and Cristian Sminchisescu.
\newblock Differentiable dynamics for articulated 3d human motion
  reconstruction.
\newblock In \emph{2022 IEEE/CVF Conference on Computer Vision and Pattern
  Recognition (CVPR)}, pages 13180--13190, 2022{\natexlab{a}}.

\bibitem[Gärtner et~al.(2022{\natexlab{b}})Gärtner, Andriluka, Xu, and
  Sminchisescu]{Gartner2022Trajectory}
Erik Gärtner, Mykhaylo Andriluka, Hongyi Xu, and Cristian Sminchisescu.
\newblock Trajectory optimization for physics-based reconstruction of 3d human
  pose from monocular video.
\newblock In \emph{2022 IEEE/CVF Conference on Computer Vision and Pattern
  Recognition (CVPR)}, pages 13096--13105, 2022{\natexlab{b}}.

\bibitem[Howell et~al.(2022)Howell, Gileadi, Tunyasuvunakool, Zakka, Erez, and
  Tassa]{howell2022predictivesampling}
Taylor Howell, Nimrod Gileadi, Saran Tunyasuvunakool, Kevin Zakka, Tom Erez,
  and Yuval Tassa.
\newblock {Predictive Sampling}: Real-time behaviour synthesis with {MuJoCo}.
\newblock \emph{arXiv preprint arXiv:2212.00541}, 2022.

\bibitem[Huang et~al.(2022)Huang, Pan, Yang, Ju, and Wang]{Huang2022MoCon}
Buzhen Huang, Liang Pan, Yuan Yang, Jingyi Ju, and Yangang Wang.
\newblock Neural mocon: Neural motion control for physically plausible human
  motion capture.
\newblock In \emph{2022 IEEE/CVF Conference on Computer Vision and Pattern
  Recognition (CVPR)}, pages 6407--6416, 2022.

\bibitem[Huang et~al.(2020)Huang, Jiang, Li, Zhang, Traish, Deng, Ferguson, and
  Da~Xu]{Huang2020endtoenddynamicmatching}
Congzhentao Huang, Shuai Jiang, Yang Li, Ziyue Zhang, Jason Traish, Chen Deng,
  Sam Ferguson, and Richard~Yi Da~Xu.
\newblock End-to-end dynamic matching network for multi-view multi-person 3d
  pose estimation.
\newblock In \emph{Computer Vision – ECCV 2020: 16th European Conference,
  Glasgow, UK, August 23–28, 2020, Proceedings, Part XXVIII}, page 477–493,
  Berlin, Heidelberg, 2020. Springer-Verlag.

\bibitem[Ingwersen et~al.(2023)Ingwersen, Mikkelstrup, Jensen, Hannemose, and
  Dahl]{ingwersen2023sportspose}
Christian~Keilstrup Ingwersen, Christian~M{\o}ller Mikkelstrup,
  Janus~N{\o}rtoft Jensen, Morten~Rieger Hannemose, and Anders~Bjorholm Dahl.
\newblock Sportspose-a dynamic 3d sports pose dataset.
\newblock In \emph{Proceedings of the IEEE/CVF Conference on Computer Vision
  and Pattern Recognition}, pages 5219--5228, 2023.

\bibitem[Isogawa et~al.(2020)Isogawa, Yuan, O'Toole, and
  Kitani]{Isogawa2020NLOS}
Mariko Isogawa, Ye Yuan, Matthew O'Toole, and Kris Kitani.
\newblock Optical non-line-of-sight physics-based 3d human pose estimation.
\newblock In \emph{2020 IEEE/CVF Conference on Computer Vision and Pattern
  Recognition (CVPR)}, pages 7011--7020, 2020.

\bibitem[Jiang et~al.(2024)Jiang, Billingham, M{\"u}ksch, Zarate, Evans,
  Oswald, Polleyfeys, Hilliges, Kaufmann, and Song]{jiang2024worldpose}
Tianjian Jiang, Johsan Billingham, Sebastian M{\"u}ksch, Juan Zarate, Nicolas
  Evans, Martin~R Oswald, Marc Polleyfeys, Otmar Hilliges, Manuel Kaufmann, and
  Jie Song.
\newblock Worldpose: A world cup dataset for global 3d human pose estimation.
\newblock In \emph{European Conference on Computer Vision}, pages 343--362.
  Springer, 2024.

\bibitem[Jin et~al.(2022)Jin, Xu, Wang, Xiao, Guo, Nie, and
  Zhao]{jin2022single}
Lei Jin, Chenyang Xu, Xiaojuan Wang, Yabo Xiao, Yandong Guo, Xuecheng Nie, and
  Jian Zhao.
\newblock Single-stage is enough: Multi-person absolute 3d pose estimation.
\newblock In \emph{CVPR}, 2022.

\bibitem[Kocabas et~al.(2019)Kocabas, Karagoz, and
  Akbas]{kocabas2019selfsupervised}
Muhammed Kocabas, Salih Karagoz, and Emre Akbas.
\newblock Self-supervised learning of 3d human pose using multi-view geometry.
\newblock In \emph{2019 IEEE/CVF Conference on Computer Vision and Pattern
  Recognition (CVPR)}, pages 1077--1086, 2019.

\bibitem[Li et~al.(2022)Li, Bian, Xu, Liu, Yu, and Lu]{li2022dd}
Jiefeng Li, Siyuan Bian, Chao Xu, Gang Liu, Gang Yu, and Cewu Lu.
\newblock D \&d: Learning human dynamics from dynamic camera.
\newblock In \emph{Computer Vision – ECCV 2022: 17th European Conference, Tel
  Aviv, Israel, October 23–27, 2022, Proceedings, Part V}, page 479–496,
  Berlin, Heidelberg, 2022. Springer-Verlag.

\bibitem[Li et~al.(2021)Li, Chen, He, Wang, Wu, and Wang]{li2021multisports}
Yixuan Li, Lei Chen, Runyu He, Zhenzhi Wang, Gangshan Wu, and Limin Wang.
\newblock Multisports: A multi-person video dataset of spatio-temporally
  localized sports actions.
\newblock In \emph{Proceedings of the IEEE/CVF International Conference on
  Computer Vision}, pages 13536--13545, 2021.

\bibitem[Liao et~al.(2024)Liao, Zhu, Wang, Hu, and
  Waslander]{liao2024mvgformer}
Ziwei Liao, Jialiang Zhu, Chunyu Wang, Han Hu, and Steven~L. Waslander.
\newblock Multiple view geometry transformers for 3d human pose estimation.
\newblock In \emph{CVPR}, 2024.

\bibitem[Lin and Lee(2021)]{lin2021multiviewplanesweep}
Jiahao Lin and Gim~Hee Lee.
\newblock { Multi-View Multi-Person 3D Pose Estimation with Plane Sweep Stereo
  }.
\newblock In \emph{2021 IEEE/CVF Conference on Computer Vision and Pattern
  Recognition (CVPR)}, pages 11881--11890, Los Alamitos, CA, USA, 2021. IEEE
  Computer Society.

\bibitem[Loper et~al.(2015)Loper, Mahmood, Romero, Pons-Moll, and
  Black]{SMPL2015loper}
Matthew Loper, Naureen Mahmood, Javier Romero, Gerard Pons-Moll, and Michael~J.
  Black.
\newblock Smpl: a skinned multi-person linear model.
\newblock \emph{ACM Trans. Graph.}, 34\penalty0 (6), 2015.

\bibitem[Luo et~al.(2022)Luo, Iwase, Yuan, and
  Kitani]{Luo2022EmbodiedSceneAware}
Zhengyi Luo, Shun Iwase, Ye Yuan, and Kris Kitani.
\newblock Embodied scene-aware human pose estimation.
\newblock In \emph{Proceedings of the 36th International Conference on Neural
  Information Processing Systems}, Red Hook, NY, USA, 2022. Curran Associates
  Inc.

\bibitem[Nocedal and Wright(2006)]{NoceWrig06}
Jorge Nocedal and Stephen~J. Wright.
\newblock \emph{Numerical Optimization}.
\newblock Springer, New York, NY, USA, 2 edition, 2006.

\bibitem[Park et~al.(2023)Park, You, Lee, and Lee]{park2023robust}
Sungchan Park, Eunyi You, Inhoe Lee, and Joonseok Lee.
\newblock Towards robust and smooth 3d multi-person pose estimation from
  monocular videos in the wild.
\newblock In \emph{ICCV}, 2023.

\bibitem[Pavlakos et~al.(2019)Pavlakos, Choutas, Ghorbani, Bolkart, Osman,
  Tzionas, and Black]{pavlakos2019expressive}
Georgios Pavlakos, Vasileios Choutas, Nima Ghorbani, Timo Bolkart, Ahmed~A.
  Osman, Dimitrios Tzionas, and Michael~J. Black.
\newblock { Expressive Body Capture: 3D Hands, Face, and Body From a Single
  Image }.
\newblock In \emph{2019 IEEE/CVF Conference on Computer Vision and Pattern
  Recognition (CVPR)}, pages 10967--10977, Los Alamitos, CA, USA, 2019. IEEE
  Computer Society.

\bibitem[Peng et~al.(2018)Peng, Abbeel, Levine, and van~de
  Panne]{peng2018deepmimic}
Xue~Bin Peng, Pieter Abbeel, Sergey Levine, and Michiel van~de Panne.
\newblock Deepmimic: example-guided deep reinforcement learning of
  physics-based character skills.
\newblock \emph{ACM Trans. Graph.}, 37\penalty0 (4), 2018.

\bibitem[Ren et~al.(2024)Ren, Xiao, and Nie]{ren2024empowering}
Ziliang Ren, Xiongjiang Xiao, and Huabei Nie.
\newblock Empowering efficient spatio-temporal learning with a 3d cnn for
  pose-based action recognition.
\newblock \emph{Sensors (Basel, Switzerland)}, 24\penalty0 (23):\penalty0 7682,
  2024.

\bibitem[Saleem et~al.(2025)Saleem, Lee, and Cai]{saleem2025biopose}
Muhammad~Usama Saleem, Gim~Hee Lee, and Yujun Cai.
\newblock Biopose: Biomechanically-accurate 3d pose estimation from monocular
  videos.
\newblock \emph{IEEE Transactions on Pattern Analysis and Machine
  Intelligence}, 2025.

\bibitem[Scott et~al.(2024)Scott, Uchida, Ding, Umemoto, Bunker, Kobayashi,
  Koyama, Onishi, Kameda, and Fujii]{scott2024teamtrack}
Atom Scott, Ikuma Uchida, Ning Ding, Rikuhei Umemoto, Rory Bunker, Ren
  Kobayashi, Takeshi Koyama, Masaki Onishi, Yoshinari Kameda, and Keisuke
  Fujii.
\newblock Teamtrack: A dataset for multi-sport multi-object tracking in
  full-pitch videos.
\newblock In \emph{Proceedings of the IEEE/CVF conference on computer vision
  and pattern recognition}, pages 3357--3366, 2024.

\bibitem[Shimada et~al.(2020)Shimada, Golyanik, Xu, and
  Theobalt]{Shimada_2020_PhysCap}
Soshi Shimada, Vladislav Golyanik, Weipeng Xu, and Christian Theobalt.
\newblock Physcap: physically plausible monocular 3d motion capture in real
  time.
\newblock \emph{ACM Trans. Graph.}, 39\penalty0 (6), 2020.

\bibitem[Srivastav et~al.(2024)Srivastav, Chen, and
  Padoy]{srivastav2024selfpose3d}
Vinkle Srivastav, Keqi Chen, and Nicolas Padoy.
\newblock Selfpose3d: Self-supervised multi-person multi-view 3d pose
  estimation.
\newblock In \emph{CVPR}, 2024.

\bibitem[Sun et~al.(2021)Sun, Bao, Liu, Fu, Black, and Mei]{sun2021romp}
Yu Sun, Qian Bao, Wu Liu, Yili Fu, Michael~J. Black, and Tao Mei.
\newblock Monocular, one-stage, regression of multiple 3d people.
\newblock In \emph{ICCV}, 2021.

\bibitem[Sweeting et~al.(2017)Sweeting, Aughey, Cormack, and
  Morgan]{sweeting2017discovering}
Alice~J Sweeting, Robert~J Aughey, Stuart~J Cormack, and Stuart Morgan.
\newblock Discovering frequently recurring movement sequences in team-sport
  athlete spatiotemporal data.
\newblock \emph{Journal of Sports Sciences}, 35\penalty0 (24):\penalty0
  2439--2445, 2017.

\bibitem[Tassa et~al.(2012)Tassa, Erez, and
  Todorov]{tassa2012trajectoryoptimization}
Yuval Tassa, Tom Erez, and Emanuel Todorov.
\newblock Synthesis and stabilization of complex behaviors through online
  trajectory optimization.
\newblock In \emph{2012 IEEE/RSJ International Conference on Intelligent Robots
  and Systems}, pages 4906--4913, 2012.

\bibitem[Todorov et~al.(2012)Todorov, Erez, and Tassa]{todorov2012mujoco}
Emanuel Todorov, Tom Erez, and Yuval Tassa.
\newblock {MuJoCo}: A physics engine for model-based control.
\newblock In \emph{2012 IEEE/RSJ International Conference on Intelligent Robots
  and Systems}, pages 5026--5033. IEEE, 2012.

\bibitem[Tripathi et~al.(2023)Tripathi, Müller, Huang, Taheri, Black, and
  Tzionas]{Tripathi2023IP}
Shashank Tripathi, Lea Müller, Chun-Hao~P. Huang, Omid Taheri, Michael~J.
  Black, and Dimitrios Tzionas.
\newblock 3d human pose estimation via intuitive physics.
\newblock In \emph{2023 IEEE/CVF Conference on Computer Vision and Pattern
  Recognition (CVPR)}, pages 4713--4725, 2023.

\bibitem[Tu et~al.(2020)Tu, Wang, and Zeng]{tu2020voxelpose}
Hanyue Tu, Chunyu Wang, and Wenjun Zeng.
\newblock Voxelpose: Towards multi-camera 3d human pose estimation in wild
  environment.
\newblock In \emph{Computer Vision – ECCV 2020: 16th European Conference,
  Glasgow, UK, August 23–28, 2020, Proceedings, Part I}, page 197–212,
  Berlin, Heidelberg, 2020. Springer-Verlag.

\bibitem[Ugrinovic et~al.(2024)Ugrinovic, Pan, Pavlakos, Paschalidou, Shen,
  Sanchez-Riera, Moreno-Noguer, and Guibas]{Ugrinovic2024MultiPhys}
Nicolas Ugrinovic, Boxiao Pan, Georgios Pavlakos, Despoina Paschalidou, Bokui
  Shen, Jordi Sanchez-Riera, Francesc Moreno-Noguer, and Leonidas Guibas.
\newblock Multiphys: Multi-person physics-aware 3d motion estimation.
\newblock In \emph{2024 IEEE/CVF Conference on Computer Vision and Pattern
  Recognition (CVPR)}, pages 2331--2340, 2024.

\bibitem[Wang et~al.(2021)Wang, Zhang, Cai, Yan, and Feng]{wang2021mvp}
Tao Wang, Jianfeng Zhang, Yujun Cai, Shuicheng Yan, and Jiashi Feng.
\newblock Direct multi-view multi-person 3d pose estimation.
\newblock In \emph{NeurIPS}, 2021.

\bibitem[Wang et~al.(2010)Wang, Li, and Zheng]{OpenCVcalib}
Y.~M. Wang, Y. Li, and J.~B. Zheng.
\newblock A camera calibration technique based on opencv.
\newblock In \emph{The 3rd International Conference on Information Sciences and
  Interaction Sciences}, pages 403--406, 2010.

\bibitem[W{\"o}hler(2009)]{wohler20093d}
Christian W{\"o}hler.
\newblock \emph{3D computer vision: efficient methods and applications}.
\newblock Springer Science \& Business Media, 2009.

\bibitem[Xu et~al.(2024)Xu, Zhang, Zhang, and Tao]{xu2022vitpose}
Yufei Xu, Jing Zhang, Qiming Zhang, and Dacheng Tao.
\newblock Vitpose++: Vision transformer for generic body pose estimation.
\newblock \emph{IEEE Trans. Pattern Anal. Mach. Intell.}, 46\penalty0
  (2):\penalty0 1212–1230, 2024.

\bibitem[Yang et~al.(2024)Yang, Odashima, Yamao, Fujimoto, Masui, and
  Jiang]{yang2023unifiedMultiView}
Fan Yang, Shigeyuki Odashima, Sosuke Yamao, Hiroaki Fujimoto, Shoichi Masui,
  and Shan Jiang.
\newblock A unified multi-view multi-person tracking framework.
\newblock \emph{Computational Visual Media}, 10\penalty0 (1):\penalty0
  137--160, 2024.

\bibitem[Ye et~al.(2023)Ye, Pavlakos, Malik, and Kanazawa]{ye2023decoupling}
Vickie Ye, Georgios Pavlakos, Jitendra Malik, and Angjoo Kanazawa.
\newblock { Decoupling Human and Camera Motion from Videos in the Wild }.
\newblock In \emph{2023 IEEE/CVF Conference on Computer Vision and Pattern
  Recognition (CVPR)}, pages 21222--21232, Los Alamitos, CA, USA, 2023. IEEE
  Computer Society.

\bibitem[Yeung et~al.(2025)Yeung, Suzuki, Tanaka, Yin, and
  Fujii]{yeung2025athletepose3d}
Calvin C.~K. Yeung, Tomohiro Suzuki, Ryota Tanaka, Zhuoer Yin, and Keisuke
  Fujii.
\newblock {AthletePose3D}: A benchmark dataset for 3d human pose estimation and
  kinematic validation in athletic movements.
\newblock \emph{arXiv preprint arXiv:2503.07499}, 2025.

\bibitem[Yi et~al.(2022)Yi, Zhou, Habermann, Shimada, Golyanik, Theobalt, and
  Xu]{Yi2022PIP}
Xinyu Yi, Yuxiao Zhou, Marc Habermann, Soshi Shimada, Vladislav Golyanik,
  Christian Theobalt, and Feng Xu.
\newblock Physical inertial poser (pip): Physics-aware real-time human motion
  tracking from sparse inertial sensors.
\newblock In \emph{2022 IEEE/CVF Conference on Computer Vision and Pattern
  Recognition (CVPR)}, pages 13157--13168, 2022.

\bibitem[Yin et~al.(2023)Yin, Guo, Kaufmann, Zarate, Song, and
  Hilliges]{yin2023hi4d}
Yifei Yin, Chen Guo, Manuel Kaufmann, Juan~Jose Zarate, Jie Song, and Otmar
  Hilliges.
\newblock Hi4d: 4d instance segmentation of close human interaction.
\newblock In \emph{2023 IEEE/CVF Conference on Computer Vision and Pattern
  Recognition (CVPR)}, pages 17016--17027, 2023.

\bibitem[Yu et~al.(2021)Yu, Park, and Lee]{RempeContactDynamics2020}
Ri Yu, Hwangpil Park, and Jehee Lee.
\newblock Human dynamics from monocular video with dynamic camera movements.
\newblock New York, NY, USA, 2021. Association for Computing Machinery.

\bibitem[Yuan et~al.(2021)Yuan, Wei, Simon, Kitani, and
  Saragih]{Yuan2021SimPoE}
Ye Yuan, Shih-En Wei, Tomas Simon, Kris Kitani, and Jason Saragih.
\newblock Simpoe: Simulated character control for 3d human pose estimation.
\newblock In \emph{2021 IEEE/CVF Conference on Computer Vision and Pattern
  Recognition (CVPR)}, pages 7155--7165, 2021.

\bibitem[Zheng et~al.(2021)Zheng, Zhu, Mendieta, Yang, Chen, and
  Ding]{zheng2021poseformer}
Ce Zheng, Sijie Zhu, Matias Mendieta, Taojiannan Yang, Chen Chen, and Zhengming
  Ding.
\newblock 3d human pose estimation with spatial and temporal transformers.
\newblock In \emph{ICCV}, 2021.

\bibitem[Zhou et~al.(2023)Zhou, Zhang, Hayder, Petersson, and
  Harandi]{zhou2023diff3dhpe}
Jieming Zhou, Tong Zhang, Zeeshan Hayder, Lars Petersson, and Mehrtash Harandi.
\newblock Diff3dhpe: A diffusion model for 3d human pose estimation.
\newblock In \emph{ICCV Workshops}, 2023.

\bibitem[Zhu et~al.(2020)Zhu, Rematas, Curless, Seitz, and
  Kemelmacher-Shlizerman]{zhu2020nba}
Luyang Zhu, Konstantinos Rematas, Brian Curless, Steven~M. Seitz, and Ira
  Kemelmacher-Shlizerman.
\newblock Reconstructing nba players.
\newblock In \emph{Proceedings of the European Conference on Computer Vision
  (ECCV)}, 2020.

\end{thebibliography}
}
\end{document}